\documentclass{article}

\usepackage{microtype}
\usepackage{graphicx}
\usepackage{multirow}
\usepackage{colortbl}
\usepackage{xcolor}
\usepackage{subfigure}
\usepackage{booktabs} % for professional tables
\usepackage{pifont}
\usepackage{footnote}

\usepackage{multirow,tabularx}
\usepackage{multicol}
\usepackage{comment}

\usepackage{verbatim}
\usepackage{tikz}
\usetikzlibrary{shapes,calc}
\tikzset{
  treenode/.style = {shape=rectangle, rounded corners,
                     draw, align=center, fill = blue!15!white}, 
  root/.style     = {rectangle, font=\scriptsize\bf, draw, align=center, fill=red!15!white},
  env/.style      = {treenode, font=\scriptsize\bf},
  env2/.style     = {treenode, draw=black,font=\scriptsize\bf ,fill=blue!15!white, very thick},
  env3/.style     = {treenode, font=\scriptsize\bf ,fill=green!15!white, very thick},
  newenv/.style     = {draw=none, font=\scriptsize\bf, align=center},
  whiteenv/.style     = {draw=none, font=\scriptsize, align=center}
}

\usepackage{hyperref}

\usepackage[accepted]{icml2019}

\newcommand{\bgdtheta}{\boldsymbol{\theta}}

\usepackage{mathtools}
\usepackage{amssymb,amsthm}

\newtheorem{corollary}{Corollary}
\newtheorem{theorem}{Theorem}

\icmltitlerunning{Task Agnostic Continual Learning Using Online Variational Bayes}

\begin{document}

\twocolumn[
\icmltitle{Task Agnostic Continual Learning Using Online Variational Bayes}



\icmlsetsymbol{equal}{*}

\begin{icmlauthorlist}
\icmlauthor{Chen Zeno}{equal,technion}
\icmlauthor{Itay Golan}{equal,technion}
\icmlauthor{Elad Hoffer}{technion}
\icmlauthor{Daniel Soudry}{technion}
\end{icmlauthorlist}

\icmlaffiliation{technion}{Technion - Israel Institute of Technology, Haifa, Israel}

\icmlcorrespondingauthor{Daniel Soudry}{daniel.soudry@gmail.com}

\icmlkeywords{Machine Learning, ICML}

\vskip 0.3in
]



\printAffiliationsAndNotice{\icmlEqualContribution} % otherwise use the standard text.

\begin{abstract}
Catastrophic forgetting is the notorious vulnerability of neural networks to the change of the data distribution while learning. This phenomenon has long been considered a major obstacle for allowing the use of learning agents in realistic continual learning settings. A large body of continual learning research assumes that task boundaries are known during training. However, research for scenarios in which task boundaries are unknown during training has been lacking. In this paper we present, for the first time, a method for preventing catastrophic forgetting (BGD) for scenarios with task boundaries that are unknown during training --- task-agnostic continual learning. Code of our algorithm is available at \href{https://github.com/igolan/bgd}{https://github.com/igolan/bgd}.
\end{abstract}

\section{Introduction}
\label{Introduction}

% Continual learning

A \emph{continual learning} algorithm is one that is faced with sequentially-arriving tasks, with no access to samples from previous or future tasks. Special measures are needed to prevent a deep neural network (DNN) from adapting only to the latest task and forgetting past knowledge --- a phenomenon known as catastrophic forgetting~\citep{mccloskey1989catastrophic}.
In this paper we present a new continual learning algorithm for training DNNs, \emph{Bayesian Gradient Descent} (BGD).
Unlike previous algorithms, our algorithm can be applied to scenarios with task boundaries that are unknown or undefined during both training and testing.

%% Paragraph2:
Various continual learning methods have been suggested in the literature. 
Due to the lack of concrete scenario definitions, many works present an inconsistent mixture of experimental results, without an explicit definition of the tested scenarios.
Recently, the authors of \cite{hsu2018re,van2018generative} identified three distinct commonly-used experimental scenarios.
They differ in the knowledge that they assume during testing, in the number of  outputs (``heads'', i.e. output dimension), and in difficulty level.
All these scenarios share a common assumption: task boundaries are known to the algorithm during training.
This is a significant disadvantage, as in many realistic applications we may not possess the knowledge about the tasks schedule --- i.e., a \textit{``task-agnostic continual learning"} setting.

To allow a concrete comparison between different methods, in section \ref{sec:Continual learning scenarios} we provide four questions for continual learning scenario definition, expanding that of~\cite{hsu2018re,van2018generative}. It yields two new scenarios where the task schedule is unknown, which were previously unaddressed.
In section \ref{sec:Experiments}, we show empirically that all scenarios are fundamentally different, and present first results on scenarios with no knowledge on tasks schedule.

% Paragraph3:
The motivation for well-defined scenarios can also be found in~\cite{farquhar2018towards}.
The authors explain why some scenarios are more interesting than others, and provide a list of desiderata for a continual learning algorithm. Our algorithm, BGD, fits most of those desiderata.
They also urge the community to focus on a specific scenario called ``class learning''. 
In section \ref{sec:Continual learning scenarios} we analyze this scenario, and present a method (``labels trick'') to improve the performance of ``class learning''. For example, on \emph{Split MNIST} it improves average accuracy from $\sim20\%$ to $\sim50\%$ for all algorithms.
This is done by understanding the information available in such scenario.

\paragraph{Main contributions}
Our main contributions are:
\vspace{-0.29cm}
\begin{enumerate}
    \item We present the first algorithm for continual learning (BGD) which is applicable to scenarios where task identity or boundaries are \emph{unknown} during both training and testing --- task-agnostic continual learning.
    \item We suggest a fine-grain categorization for continual learning scenarios to allow well-defined comparison.
    \item We present the ``labels trick'', a way to improve performance in the ``class learning'' scenario.
\end{enumerate}
\vspace{-0.29cm}

\section{Continual learning scenarios}
\label{sec:Continual learning scenarios}

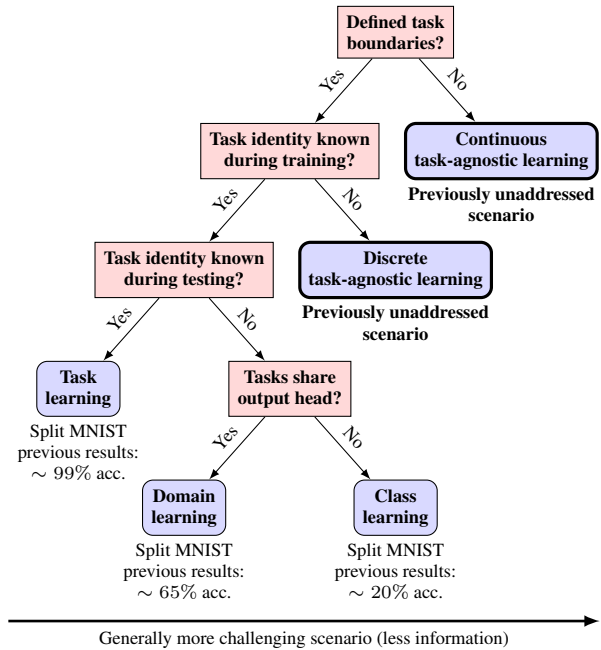
\begin{figure}[t]
\centering
\begin{tikzpicture}
  [
    grow                    = down,
    sibling distance        = 8em,
    level distance          = 4.5em,
    edge from parent/.style = {draw, -latex},
    every node/.style       = {font=\scriptsize},
    sloped,
    >=latex
  ]
  \node [root] {Defined task \\ boundaries?}
    child { node [root] {Task identity known \\ during training?}
      child { node [root] {Task identity known \\ during testing?}
        child { node [root]  [env] {Task \\learning}
          child[level distance=2.4em] { node (TL) [whiteenv] {Split MNIST \\ previous results: \\ $\sim99\%$ acc.} edge from parent[draw=none] }
          edge from parent node [above] {Yes}}
        child { node [root] {Tasks share \\ output head?}
            child { node [env] {Domain \\learning}
              child[level distance=2.4em] { node [whiteenv] {Split MNIST \\ previous results: \\ $\sim65\%$ acc.} edge from parent[draw=none] }
                edge from parent node [above] {Yes}}
             child { node [env] {Class \\learning}
               child[level distance=2.4em] { node [whiteenv] {Split MNIST \\ previous results: \\ $\sim20\%$ acc.} edge from parent[draw=none] }
                edge from parent node [above] {No}}
          edge from parent node [above] {No} }
          edge from parent node [above] {Yes}}
      child { node (TAL) [env2]  {Discrete \\ task-agnostic learning}
      child[level distance=2em] { node [newenv] {Previously unaddressed \\ scenario} edge from parent[draw=none] }
              edge from parent node [above] {No}}
        edge from parent node [above] {Yes} }
        child { node [env2] {Continuous \\ task-agnostic learning}
        child[level distance=2em] { node [newenv] {Previously unaddressed \\ scenario} edge from parent[draw=none] }
      edge from parent node [above] {No} };
    \path[draw,line width=1,->] ($(TL.south west)-(0,1.75cm)$) -- node[midway,below]{Generally more challenging scenario (less information)} ($(TL.south east)+(6cm,-1.75cm)$); 
\end{tikzpicture}
\caption{Continual learning scenario characterization. Each scenario is fundamentally different. This paper presents first results on task-agnostic scenarios. Previous results are average accuracy over all tasks as reported in \cite{hsu2018re}.}
\label{fig:Continual learning scenarios}
\end{figure}

In recent works,~\cite{hsu2018re,van2018generative} defined three continual learning scenarios: ``task learning'', ``domain learning'' and ''class learning''.
We review these scenarios, denoting $T$ as the number of tasks, and $C$ as the number of classes per task.\footnote{We assume all tasks have the same number of classes for simplicity, but this assumption may be omitted.}
The term ``head'' is used for the DNN output (the last linear classifier).
%, and ``active heads'' are the outputs which the network is trained on or the output which are used for inference.
Knowledge about task identity means that a sample $x_{i}$ is tagged with task number $t_{i}$, alongside the label $y_{i}$: $(x_{i}, y_{i}, t_{i})$.

\paragraph{Previous scenario definitions}
In task, domain, and class learning, task boundaries are well-defined, and task identity is \textit{known} during \textit{training}.
The difference arises from knowledge on task identity during \textit{testing}, and the number of heads.
In task learning, the task identity is \textit{known} during testing. The best results are achieved by assigning a different set of heads for each task ($T\times C$ outputs), and using only the $C$ heads of the current task during training and testing.
In both domain and class learning, the task identity is \textit{unknown} during testing. Therefore, we cannot easily choose the correct set of heads on testing.
In domain learning, a single set of heads is shared among all tasks ($C$ outputs). While in class learning a different set of heads represents each task classes ($T\times C$ outputs).
Specifically, in class learning, during testing one must use all $T\times C$ heads, which includes heads of other tasks, therefore the network needs to infer not only the correct label, but also the correct task.

\begin{figure}[h]
\begin{center}
\centerline{\includegraphics[width= 0.8\columnwidth]{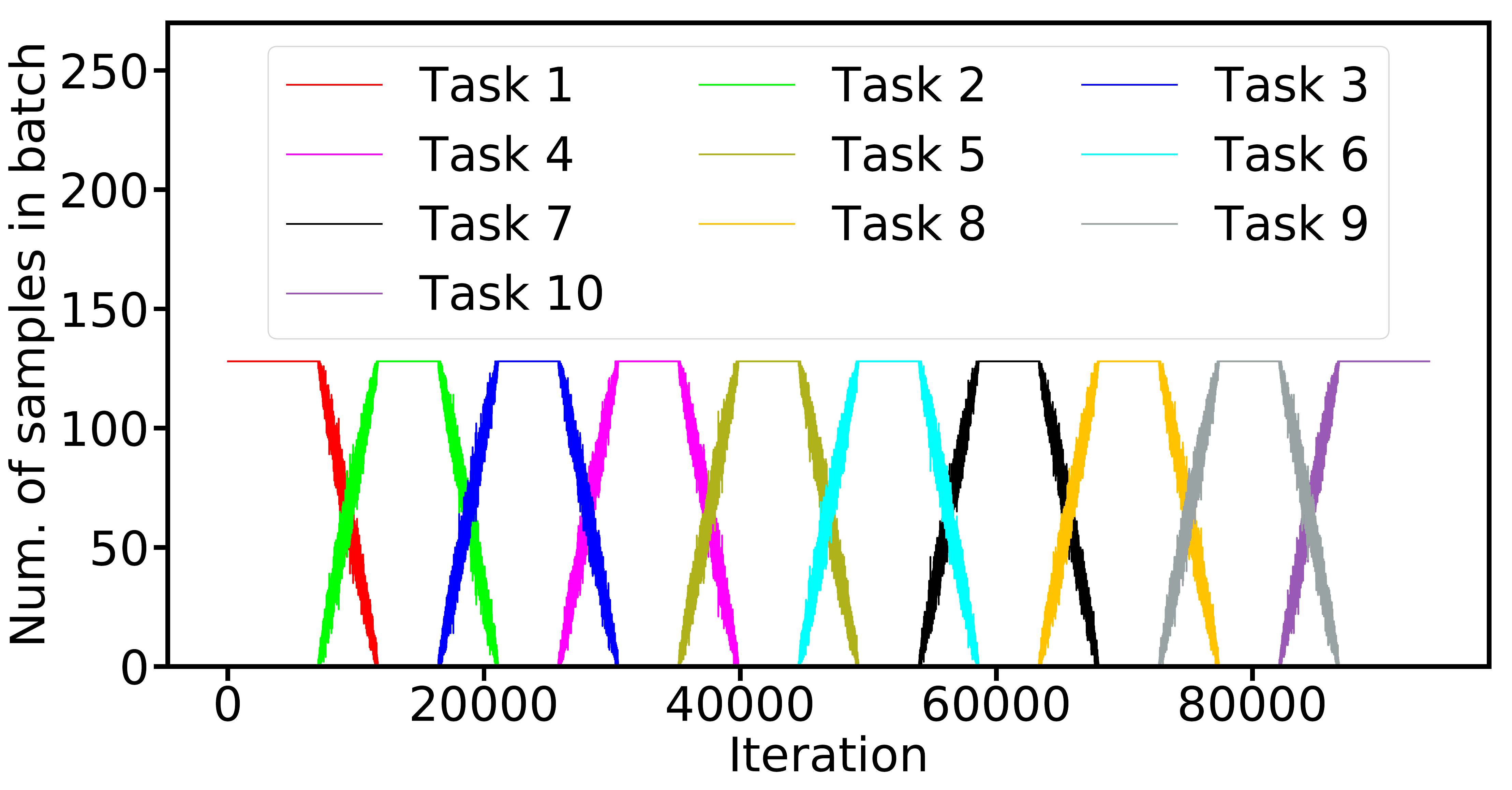}}
\caption{Distribution of samples from each task as a function of iteration. The tasks are not changed abruptly, but slowly over time --- tasks boundaries are undefined. Moreover, the algorithm has no access to this distribution. Here, number of samples from each task in each batch is a random variable drawn from a distribution over tasks, and this distribution changes over time (iterations).}
\label{fig:tasks_distribution}
\vspace{-1em}
\end{center}
\end{figure}

\paragraph{Expanding scenario definitions}
We expand those definitions, and suggest a fine-grain categorization of continual learning scenarios. 
To characterize a scenario, one should answer four questions:
\vspace{-0.29cm}
\begin{enumerate}
    \setlength{\parskip}{0pt} \setlength{\itemsep}{0pt plus 2pt}
    \item Are the boundaries between tasks well defined?
    \item Is the task identity known during training?
    \item Is the task identity known during inference (test)?
    \item Are tasks share the output head? ($C$ vs. $T\times C$ outputs).
\end{enumerate}
\vspace{-0.29cm}
Additional questions that affect performance are:
\vspace{-0.29cm}
\begin{enumerate}
    \setcounter{enumi}{4}
    \setlength{\parskip}{0pt} \setlength{\itemsep}{0pt plus 2pt}
    \item For how long each task is trained?\footnote{When task boundaries are undefined we can not define an epoch, as the notion of ``transition over the whole data set'' is undefined.}
    \item Is the number of tasks known? How many tasks exist?
    \item General questions: network architecture, optimizer etc.
\end{enumerate}
\vspace{-0.29cm}
The tasks boundaries can be undefined in two different ways: (A) The transition between tasks occurs slowly over time, so the algorithm gets a mixture of samples from two different tasks during the transition (Figure \ref{fig:tasks_distribution}); (B) The data itself changes continuously towards a new distribution (for example, images of cheetahs that slowly change into cats).
In all task-agnostic scenarios, the algorithm does not have any kind of knowledge on the distribution over the tasks.
Figure \ref{fig:Continual learning scenarios} summarizes scenario characterization as a decision tree.

\paragraph{Task agnostic continual learning}
Current methods for continual learning require to be informed when a task is switched during training in order to take some core action (e.g., changing parameters in the loss function).
When this knowledge is unavailable, such as in task-agnostic scenarios, those algorithms are inapplicable (see discussion in subsection \ref{subsection:exp_cont_permuted}).
Unlike previous approaches, BGD can be applied without any modification to scenarios where the task boundaries are \textit{unknown} or \textit{undefined} at all.

\paragraph{Labels trick}
In class learning, the output heads are not shared, so different tasks do not share the same label space. In such case, we can infer the task identity during training even if it is not provided explicitly, solely by inspecting the labels of the current batch (as different tasks will have different labels due to the non-shared heads).
The basic idea is:
\vspace{-0.29cm}
\begin{center}
    \emph{Heads are not shared $\Rightarrow$ Task identity known in training.}
\end{center}
\vspace{-0.29cm}
If task identity is known during training, we can train only the heads of the current task.
Using this notion, we suggest the ``labels trick'' for class learning.
The trick is training only the heads which correspond to labels that exist in the current batch (i.e. heads of the current task), in contrast to the common practice of training all heads in such scenario.
For example, if a batch consists of samples only with labels $(3,4)$, we calculate the loss only over the heads corresponding to $(3,4)$, instead of over all heads $(0,1,2,3,4...T\times C-1)$.
This way, we are able to train the network as in ``task learning''. The error stems only from wrong task prediction during testing, since task identity is unavailable during testing.
This trick improves class learning results from $\sim20\%$ to $\sim50\%$ (see subsection \ref{subsection:labels_trick_experiments}). Note that the scenario setup is kept the same, and we do not use any new information --- the labels were available during training also in previous papers results.

\section{Related work} 
\label{Related work}
\paragraph{Approaches for continual learning}
Approaches for continual learning can be generally divided into four main categories: (1) Architectural approach; (2) Rehearsal approach; (3) Regularization approach; (4) Bayesian approach. 

\textit{Architectural approaches} alter the network architecture to adapt to new tasks (e.g. \cite{rusu2016progressive}).
\textit{Rehearsal approaches} use memory\footnote{Rehearsal approaches include GANs which can be used as a sort of memory.} to allow re-training on examples from previous tasks (e.g. \cite{shin2017continual}).
\textit{Regularization approaches} use some penalty on deviations from previous task weights.
\textit{Bayesian approaches} use Bayes' rule with previous task posterior as the current prior.
Each approach has pros and cons. For example, architectural approaches may result in extremely big architecture for a large number of tasks, while rehearsal approaches may be problematic in terms of privacy.
For an extensive review on continual learning methods see \cite{parisi2018continual}.
This paper focuses on \textit{regularization and Bayesian approaches} -- the architecture is fixed, and no memory is used to retrain on previous tasks.

In \emph{regularization} approaches, a regularization term is added to the loss function. ``Elastic weight consolidation'' (EWC) proposed by~\cite{kirkpatrick2017overcoming}, slows the conversion of parameters important to the previous tasks by adding a quadratic penalty on the difference between the optimal parameters of the previous task and the current parameters. The importance of each parameter is measured using the diagonal of the Fisher information matrix of the previous tasks.
``Synaptic Intelligence'' (SI) proposed by~\cite{zenke2017continual} also uses a quadratic penalty, %on the difference between the optimal parameter of the previous task and the current parameter. 
however, the importance is measured by the path length of the updates on the previous task. \cite{chaudhry2018riemannian} proposed a generalization of the online version of EWC and modified version of SI to achieve better performance. ``Memory Aware Synapses'' (MAS) proposed by~\cite{aljundi2018memory} also uses a quadratic penalty, however, the importance is measured by the sensitivity of the output function.
``Learning without Forgetting'' (LwF) proposed by~\cite{li2017learning} uses knowledge distillation so the network outputs of the new task enforced to be similar to the network of the previous tasks.

The \textit{Bayesian approach} provides a solution to continual learning in the form of Bayes' rule. When data arrives sequentially, the posterior distribution of the parameters for the previous task is used as a prior for the new task. ``Variational Continual Learning'' (VCL) proposed by~\cite{nguyen2017variational} used online variational inference combined with the standard variational Bayes approach developed by~\cite{blundell2015weight}, ``Bayes By Backprop'' (BBB) to reduce catastrophic forgetting by replacing the prior on a task switch. In the BBB approach, a ``mean-field approximation'' is applied, assuming weights are independent of each other (the covariance matrix is diagonal).
%meaning that a fully factorized distribution is used to approximate the posterior distribution. Thus, VCL is a diagonal method.
\cite{ritter2018online} suggest using Bayesian online learning with a Kronecker factored Laplace approximation to attain a non-diagonal method for reducing catastrophic forgetting, which allows the algorithm to take into account interactions between weights within the same layer.

\paragraph{Bayesian Neural Networks}
In this work, we use the Bayesian approach for continual learning, specifically we use the online version of variational Bayes. Next we present a review on Bayesian neural networks and variational Bayes methods.
Bayesian inference for neural networks has been a subject of interest over many years. As exact Bayesian inference is intractable (for any realistic network size), much research has been focused on approximation techniques. Most modern techniques stemmed from previous seminal works which used either a Laplace approximation~\citep{Mackay1992}, variational methods~\citep{Hinton1993}, or Monte Carlo methods~\citep{Neal94}. In the last years, many methods for approximating the posterior distribution have been suggested, falling into one of these categories. Those methods include assumed density filtering~\citep{soudry2014expectation,hernandez2015probabilistic}, approximate power Expectation Propagation~\citep{hernandez2016black}, Stochastic Langevin Gradient Descent~\citep{welling2011bayesian,balan2015bayesian}, incremental moment matching~\citep{Lee2017} and variational Bayes~\citep{graves2011practical,blundell2015weight}.

Practical variational Bayes for modern neural networks was first introduced by~\cite{graves2011practical}, where parametric distribution is used to approximate the posterior distribution by minimizing the variational free energy. Calculating the variational free energy is intractable for general neural networks, and thus~\cite{graves2011practical}
estimated its gradients using a biased Monte Carlo method, and used stochastic gradient descent (SGD)
to perform minimization. In a later work,~\cite{blundell2015weight} used a re-parameterization trick to introduce an unbiased estimator for the gradients. Variational Bayes methods were also used extensively on various probabilistic models including recurrent neural networks~\citep{graves2011practical}, auto-encoder~\citep{kingma2013auto} and fully connected networks~\citep{blundell2015weight}. \cite{martens2015optimizing,zhang2017noisy,khan2018fast} suggested using the connection between natural gradient descent~\citep{amari1998natural} and
variational inference to perform natural gradient optimization in deep neural networks.

\paragraph{Major differences} The major differences between BGD and previous works are as follows:
\vspace{-0.29cm}
\begin{enumerate}
    \item  \textbf{Task agnostic scenarios:} BGD (our method) is a task-agnostic algorithm --- it  does not require any information on task identity. In contrast to previous methods (e.g. EWC, SI, VCL) which are based on some core action taken on task switch. 
    \item  \textbf{Closed-form update rule:} We update the posterior over the weights in closed-form, in contrast to BBB (and VCL which relies on BBB) which use SGD optimizer on the variational free energy.
\end{enumerate}
\vspace{-0.29cm}

\section{Theory}
\label{sec:Theory}
The process of Bayesian inference requires a full probability model
providing a joint probability distribution over the data and the model
parameters. The joint probability distribution can be written as a
product of two distributions:
\begin{equation}
p\left(D,\bgdtheta\right)=p\left(D|\bgdtheta\right)p\left(\bgdtheta\right)\,,
\end{equation}
where $p\left(D|\bgdtheta\right)$ is the likelihood function of the data
set $D$, and $p\left(\bgdtheta\right)$ is the prior distribution of
the parameters $\bgdtheta$. The posterior distribution can be calculated
using Bayes' rule:
\begin{equation}
p\left(\bgdtheta|D\right)=\frac{p\left(D|\bgdtheta\right)p\left(\bgdtheta\right)}{p\left(D\right)} \,,
\end{equation}
where $p\left(D\right)$ is calculated using the sum rule. 

In this paper we will focus on the online version of Bayesian inference, in which the data arrives sequentially, and we do a sequential update of the posterior distribution each time that new data arrives. In each step, the previous posterior distribution is used as the new prior distribution. Therefore, according to Bayes' rule, the posterior distribution at time $n$ is given by:
\begin{equation}
p\left(\bgdtheta|D_{n}\right)=\frac{p\left(D_{n}|\bgdtheta\right)p\left(\bgdtheta|D_{n-1}\right)}{p\left(D_{n}\right)} \,. \label{eq:online bayes}
\end{equation}
Unfortunately, calculating the posterior distribution is intractable
for most practical probability models;
therefore, we approximate the true posterior using variational methods.

\subsection{Online variational Bayes}

In variational Bayes \citep{graves2011practical}, a parametric distribution $q\left(\bgdtheta|\phi\right)$ 
is used for approximating the true posterior distribution $p\left(\bgdtheta\right|D)$
by minimizing the Kullback-Leibler (KL) divergence with the true posterior
distribution.
\begin{equation}
\mathrm{\mathrm{KL}}\left(q\left(\bgdtheta|\phi\right)||p\left(\bgdtheta|D\right)\right)=-\mathbb{E}_{\bgdtheta\sim q\left(\bgdtheta|\phi\right)}\left[\log\frac{p\left(\bgdtheta|D\right)}{q\left(\bgdtheta|\phi\right)}\right]
\end{equation}
The optimal variational parameters are the solution of the following optimization problem:
\begin{align}
\phi^{*} & =\arg\min_{\phi}\int q\left(\bgdtheta|\phi\right)\log\frac{q\left(\bgdtheta|\phi\right)}{p\left(\bgdtheta|D\right)}d\bgdtheta\nonumber \\
& =\arg\min_{\phi}\int q\left(\bgdtheta|\phi\right)\log\frac{q\left(\bgdtheta|\phi\right)}{p\left(D|\bgdtheta\right)p\left(\bgdtheta\right)}d\bgdtheta\nonumber \\
& =\arg\min_{\phi}\mathbb{E}_{\bgdtheta\sim q\left(\bgdtheta|\phi\right)}\left[\log\left(q\left(\bgdtheta|\phi\right)\right)\right. 
\nonumber \\  & -\log\left(p\left(\bgdtheta\right)\right) +\mathrm{L}\left(\bgdtheta\right)\left.\right],
\end{align}
where $\mathrm{L}\left(\bgdtheta\right)=-\log \left(p\left(D|\bgdtheta\right)\right)$
is the log-likelihood cost function.\footnote{Note that we define a cumulative log-likelihood cost function over the data.}

In online variational Bayes \citep{broderick2013streaming}, we aim to find the posterior in an online
setting, where the data arrives sequentially. Similar to Bayesian
inference we use the previous approximated posterior as the new prior
distribution. For example, at time $n$ the optimal variational
parameters are the solution of the following optimization problem:
\begin{align}
\phi^{*} & =\arg\min_{\phi}\int q_{n}\left(\bgdtheta|\phi\right)\log\frac{q_{n}\left(\bgdtheta|\phi\right)}{p\left(\bgdtheta|D_{n}\right)}d\bgdtheta\nonumber \\ & = \arg\min_{\phi}\int q_{n}\left(\bgdtheta|\phi\right)\log\frac{q_{n}\left(\bgdtheta|\phi\right)}{p\left(D_{n}|\bgdtheta\right)q_{n-1}\left(\bgdtheta\right)}d\bgdtheta\nonumber \\
 & =\arg\min_{\phi}\mathbb{E}_{\bgdtheta\sim q_{n}\left(\bgdtheta|\phi\right)}\left[\log\left(q_{n}\left(\bgdtheta|\phi\right)\right)\right.  
 \nonumber \\ &-\log\left(q_{n-1}\left(\bgdtheta\right)\right)+\mathrm{L_{n}\left(\bgdtheta\right)}\left.\right], \label{eq:optimization - online}
\end{align}
where $\mathrm{L}_n\left(\bgdtheta\right)=-\log \left(p \left(D_{n}|\bgdtheta\right)\right)$
is the log-likelihood cost function.

\subsection{Diagonal Gaussian approximation}

A standard approach is to define the parametric distribution $q\left(\bgdtheta|\phi\right)$ so that all the components of the parameter vector $\theta$ would be factorized, i.e. independent (``mean-field approximation''). In addition, in this paper we will focus on the case
in which the parametric distribution $q\left(\bgdtheta|\phi\right)$ and
the prior distribution are Gaussian. Therefore:
\begin{align}
q_{n}\left(\bgdtheta|\phi\right) &=\prod_{i}\mathcal{N}\left(\theta_{i}|\mu_{i},\sigma_{i}^{2}\right) \nonumber \\ q_{n-1}\left(\bgdtheta\right)&=\prod_{i}\mathcal{N}\left(\theta_{i}|m_{i},v_{i}^{2}\right)\,. \label{eq:prior and posterior}
\end{align}

In order to solve the optimization problem in \eqref{eq:optimization - online}, we use
the unbiased Monte Carlo gradients, similarly to~\cite{blundell2015weight}. We define a deterministic transformation:
\begin{align}
\theta_{i} &=\mu_{i}+\varepsilon_{i}\sigma_{i} \nonumber \\
\varepsilon_{i} &\sim\mathcal{N}\left(0,1\right) \nonumber \\
\phi &=\left(\boldsymbol{\mu},\boldsymbol{\sigma}\right)\,. \label{eq:reparametrization}
\end{align}
The following holds:
\begin{equation}
\frac{\partial}{\partial\phi}\mathbb{E}_{\bgdtheta}\left[f\left(\bgdtheta,\phi\right)\right]=\mathbb{E}_{\varepsilon}\left[\frac{\partial f\left(\bgdtheta,\phi\right)}{\partial\bgdtheta}\frac{\partial\bgdtheta}{\partial\phi}+\frac{\partial f\left(\bgdtheta,\phi\right)}{\partial\phi}\right],
\end{equation}
where $f\left(\bgdtheta,\phi\right)=\log\left(q_{n}\left(\bgdtheta|\phi\right)\right)-\log\left(q_{n-1}\left(\bgdtheta\right)\right)+\mathrm{L\left(\bgdtheta\right)}$.
Therefore, we can find a critical point of the objective function
by solving the following set of equations:
\begin{equation}
\mathbb{E}_{\varepsilon}\left[\frac{\partial f\left(\bgdtheta,\phi\right)}{\partial\bgdtheta}\frac{\partial\bgdtheta}{\partial\phi}+\frac{\partial f\left(\bgdtheta,\phi\right)}{\partial\phi}\right]=0\,. \label{eq:critical point}
\end{equation}
Substituting \eqref{eq:prior and posterior} and \eqref{eq:reparametrization}
into \eqref{eq:critical point} we obtain (see Appendix \ref{appendix:Derivation of eqs} for additional
details):
\begin{equation}
\mu_{i}=m_{i}-v_{i}^{2}\mathbb{E}_{\varepsilon}\left[\frac{\partial \mathrm{L}_n\left(\bgdtheta\right)}{\partial\theta_{i}}\right]\,, \label{eq:mu implicit solution}
\end{equation}
\begin{align}
\sigma_{i}&=v_{i}\sqrt{\!1+\!\left(\frac{1}{2}v_{i}\mathbb{E}_{\varepsilon}\left[\frac{\partial \mathrm{L}_n\left(\bgdtheta\right)}{\partial\theta_{i}}\varepsilon_{i}\right]\right)^{2}} \nonumber \\ &-\frac{1}{2}v_{i}^{2}\mathbb{E}_{\varepsilon}\left[\frac{\partial \mathrm{L}_n\left(\bgdtheta\right)}{\partial\theta_{i}}\varepsilon_{i}\right]\,. \label{eq:sigma implicit solution}
\end{align}
Note that \eqref{eq:mu implicit solution} and \eqref{eq:sigma implicit solution} are implicit equations,
since the derivative $\frac{\partial \mathrm{L}_n\left(\bgdtheta\right)}{\partial\theta_{i}}$
is a function of $\mu_{i}$ and $\sigma_{i}$.
We approximate the solution using a single explicit iteration of this equation, i.e. evaluate the derivative $\frac{\partial \mathrm{L}_n\left(\bgdtheta\right)}{\partial\theta_{i}}$ using the prior parameters.

\paragraph{BGD algoirthm}
The approximation above results in an explicit closed-form update rule for $\mu$ and $\sigma$.
This approximation becomes more accurate when we are near the solution of \eqref{eq:mu implicit solution}. If we are not near the solution, this can lead to a slow rate of convergence. To compensate for this we added a "learning rate" hyper-parameter $\eta$ to adjust the convergence rate.  
Also, note that in the theoretical derivation we assumed that the data arrives sequentially, in an online setting. In practice, however, the log-likelihood cost function does not converge after one epoch since we use a single explicit iteration. Therefore, in Algorithm~\ref{algo:bgd} we repeatedly go over the training set until the convergence criterion is met.
In addition, the expectations are approximated using Monte Carlo sampling method (we use $K$ Monte Carlo samples). Those Monte Carlo samples are the main difference from SGD in terms of computation complexity, see Appendix \ref{appendix:complexity discussion} for complexity analysis.
The full algorithm is described in Algorithm \ref{algo:bgd}.

\begin{algorithm}[ht]
\caption{Bayesian Gradient Descent (BGD)}
\begin{description}
\item [{Initialize}] $\mu,\sigma,\eta, K$
\item [{Repeat}]~
\begin{description}
\item [{$\!\!\!\!\!\!\!\!\!\!\!\! \mu_{i}\leftarrow\mu_{i}-\eta\sigma_{i}^{2}\mathbb{E}_{\varepsilon}\left[\frac{\partial \mathrm{L}_n\left(\bgdtheta\right)}{\partial\theta_{i}}\right]$}]~
\item [{$\!\!\!\!\!\!\!\!\!\!\!\!
\sigma_{i}\leftarrow\sigma_{i}\sqrt{1+\left(\frac{1}{2}\sigma_{i}\mathbb{E}_{\varepsilon}\left[\frac{\partial \mathrm{L}_n\left(\bgdtheta\right)}{\partial\theta_{i}}\varepsilon_{i}\right]\right)^{2}}-\frac{1}{2}\sigma_{i}^{2}\mathbb{E}_{\varepsilon}\left[\frac{\partial \mathrm{L}_n\left(\bgdtheta\right)}{\partial\theta_{i}}\varepsilon_{i}\right]$}]~
\end{description}
\item [{Until}] convergence criterion is met.
\label{algo:bgd}
\end{description}
The expectations are estimated using Monte Carlo method, with $\theta_{i}^{(k)}=\mu_{i}+\varepsilon_{i}^{(k)}\sigma_{i}$:
\[
\mathbb{E}_{\varepsilon}\left[\frac{\partial \mathrm{L}_n\left(\bgdtheta\right)}{\partial\theta_{i}}\right]\approx\frac{1}{K}\sum_{k=1}^{K}\frac{\partial \mathrm{L}_n\left(\bgdtheta^{(k)}\right)}{\partial\theta_{i}}
\]
\[
\mathbb{E}_{\varepsilon}\left[\frac{\partial \mathrm{L}_n\left(\bgdtheta\right)}{\partial\theta_{i}}\varepsilon_{i}\right]\approx\frac{1}{K}\sum_{k=1}^{K}\frac{\partial \mathrm{L}_n\left(\bgdtheta^{(k)}\right)}{\partial\theta_{i}}\varepsilon_{i}^{(k)}
\]
\end{algorithm}

\subsection{Theoretical properties of Bayesian Gradient Descent}
\label{theoretical_properties}
Bayesian Gradient Descent (BGD) consists of a gradient descent algorithm
for $\mu$, and a recursive update rule for $\sigma$, both result from an approximation to the online Bayes update in \eqref{eq:online bayes}, hence the name BGD. The learning rate of $\mu_{i}$ is proportional to the uncertainty in the parameter $\theta_{i}$ according to the prior distribution. During the learning process, as more data is seen, the learning rate decreases for parameters with a high degree of certainty, while the learning rate increases for parameters with a high degree of uncertainty. Next, we establish this intuitive idea more precisely.

It is easy to verify that the update rule for $\sigma$ is a strictly monotonically decreasing
function of $\mathbb{E}_{\varepsilon}\left[\frac{\partial \mathrm{L}_n\left(\bgdtheta\right)}{\partial\theta_{i}}\varepsilon_{i}\right]$.
Therefore:
\begin{align} 
\mathrm{E}_{\varepsilon}\left[\frac{\partial L\left(\bgdtheta\right)}{\partial\theta_{i}}\varepsilon_{i}\right] & >0\Longrightarrow\sigma_{i}\left(n\right)<\sigma_{i}\left(n-1\right)\nonumber \\
\mathrm{E}_{\varepsilon}\left[\frac{\partial L\left(\bgdtheta\right)}{\partial\theta_{i}}\varepsilon_{i}\right] & <0\Longrightarrow\sigma_{i}\left(n\right)>\sigma_{i}\left(n-1\right)\nonumber \\
\mathrm{E}_{\varepsilon}\left[\frac{\partial L\left(\bgdtheta\right)}{\partial\theta_{i}}\varepsilon_{i}\right] & =0\Longrightarrow\sigma_{i}\left(n\right)=\sigma_{i}\left(n-1\right) \,.
\end{align}
Next, using a Taylor expansion, we show that for small values of $\sigma$, the quantity $\!\mathbb{E}_{\varepsilon}\!\left[\!\frac{\partial \mathrm{L}_n\left(\bgdtheta\right)}{\partial\theta_{i}}\varepsilon_{i}\right]$, approximates the curvature of the loss:
\begin{align*}
 & \!\mathbb{E}_{\epsilon}\!\left[\!\frac{\partial \mathrm{L}_n\left(\bgdtheta\right)}{\partial\theta_{i}}\varepsilon_{i}\right]\! \\ & = 
 \mathbb{E}_{\varepsilon}\!\left[\left(\frac{\partial \mathrm{L}_n\left(\boldsymbol{\mu}\right)}{\partial\theta_{i}}+\sum_{j}\frac{\partial^{2}\mathrm{L}_n\left(\boldsymbol{\mu}\right)}{\partial\theta_{i}\partial\theta_{j}}\epsilon_{j}\sigma_{j}+O\left(\left\Vert \boldsymbol{\sigma}\right\Vert ^{2}\right)\right)\varepsilon_{i}\right]\! \\
 & =\frac{\partial^{2}\mathrm{L}_n\left(\boldsymbol{\mu}\right)}{\partial^{2}\theta_{i}}\sigma_{i}\!+\!O\left(\left\Vert \boldsymbol{\sigma}\right\Vert ^{2}\right)\,,
\end{align*}
where we used $\mathbb{E}_{\varepsilon}\left[\varepsilon_{i}\right]=0$ and $\mathbb{E}_{\varepsilon}\left[\varepsilon_{i}\varepsilon_{j}\right]=\delta_{ij}$ in the last line. Thus, in this case, $\mathbb{E}_{\varepsilon}\left[\frac{\partial \mathrm{L}_n\left(\bgdtheta\right)}{\partial\theta_{i}}\varepsilon_{i}\right]$ is a finite difference approximation to the component-wise product of the diagonal of the Hessian of the loss, and the vector $\sigma$. Therefore, we expect that the uncertainty (learning rate) would decrease in areas with positive curvature (e.g., near local minima), or increase in areas with high negative curvature (e.g., near maxima, or saddles). This seems like a ``sensible'' behavior of the algorithm, since we wish to converge to local minima, and escape saddles. This is in contrast to many common optimization methods, which are either insensitive to the sign of the curvature, or use it the wrong way~\citep{Dauphin2014}.

In the case of strongly convex loss, we can make a more rigorous statement, which we prove in Appendix \ref{appendix:Proof of Theorem}.

\begin{theorem} \label{theorem}
We examine BGD with a diagonal Gaussian distribution for $\bgdtheta$. If ${\mathrm{L}_n}\left(\bgdtheta\right)$ is a strongly convex function with parameter $m_{n}>0$ and a continuously
differentiable function over $\mathbb{R}^{n}$, then 
$\mathbf{\mathbb{E}}_{\varepsilon}\left[\frac{\partial \mathrm{L}_n\left(\bgdtheta\right)}{\partial\theta_{i}}\varepsilon_{i}\right]\geq m_n\sigma_i>0$.
\end{theorem}

\begin{corollary} \label{corollary}
 If $\mathrm{L}_n\left(\bgdtheta\right)$ is strongly convex (concave) function for all $n\in \mathbb{N} $, then the sequence $\left\{\sigma_{i}(n)\right\}_{n=1}^{\infty}$
is strictly monotonically decreasing (increasing).
\end{corollary}

Furthermore, one can generalize these results and show that if a restriction of $\mathrm{L}_n\left(\bgdtheta\right)$ to an axis $\theta_{i}$ is strongly convex (concave) for all $n\in \mathbb{N}$, then $\left\{\sigma_{i}(n)\right\}_{n=1}^{\infty}$
is monotonic decreasing (increasing).

Therefore, in the case of a strongly convex loss function, $\sigma_{i}=0$ is the only stable point of \eqref{eq:sigma implicit solution}, which means that we collapse to a point estimation similar to SGD. However, for neural networks $\sigma_{i}$ does not generally converge to zero. In this case, the stable point $\sigma_{i}=0$ is generally not unique, since $\mathbb{E}_{\varepsilon}\left[\frac{\partial \mathrm{L}\left(\bgdtheta\right)}{\partial\theta_{i}}\varepsilon_{i}\right] $ implicitly depends on $\sigma_{i}$. 

In Appendix \ref{appendix:5000 epochs training} we show the histogram of STD values on MNIST when training for 5000 epochs and demonstrate we do not collapse to a point estimation.

\paragraph{BGD in continual learning} In the case of over-parameterized models and continual learning, only a part of the weights is essential for each task. We hypothesize that if a weight $\theta_{i}$ is important to the current task, this implies that, near the minimum, the function  $\mathrm{L}_i=\mathrm{L}(\bgdtheta)|_{(\bgdtheta)_i = \theta_i}$  is locally convex. Corollary \ref{corollary} suggests that in this case $\sigma_{i}$ would be small. In contrast, the loss will have a flat curvature in the direction of weights which are not important to the task. Therefore, these unimportant weights may have a large uncertainty $\sigma_{i}$.
Since BGD introduces the linkage between learning rate and the uncertainty (STD), the training trajectories in the next task would be restricted along the less important weights leading to a good performance on the new task, while retaining the performance on the current task. The use of BGD to continual learning exploits the inherent features of the algorithm, and it does not need any explicit information on tasks --- it is completely unaware of the notion of tasks.

\section{Experiments}
\label{sec:Experiments}
We show various empirical results to support the claims made throughout this paper.
Description of the used datasets can be found in Appendix \ref{appendix:Datasets}, and full implementation details can be found in Appendix \ref{appendix:Implementation details}. In all experiments, we conducted a hyper-parameter search. The code used for BGD experiments is available at \href{https://github.com/igolan/bgd}{https://github.com/igolan/bgd}.
We aimed to provide as a broad variety of comparisons as possible, however, due to the lack of open-source implementations / architecture-specific implementations of other algorithms, not all algorithms are evaluated in all experiments.

\subsection{Continuous task agnostic scenario}
\label{subsection:exp_cont_permuted}
\begin{figure}[h]
\begin{center}
\centerline{\includegraphics[width= 0.8\columnwidth]{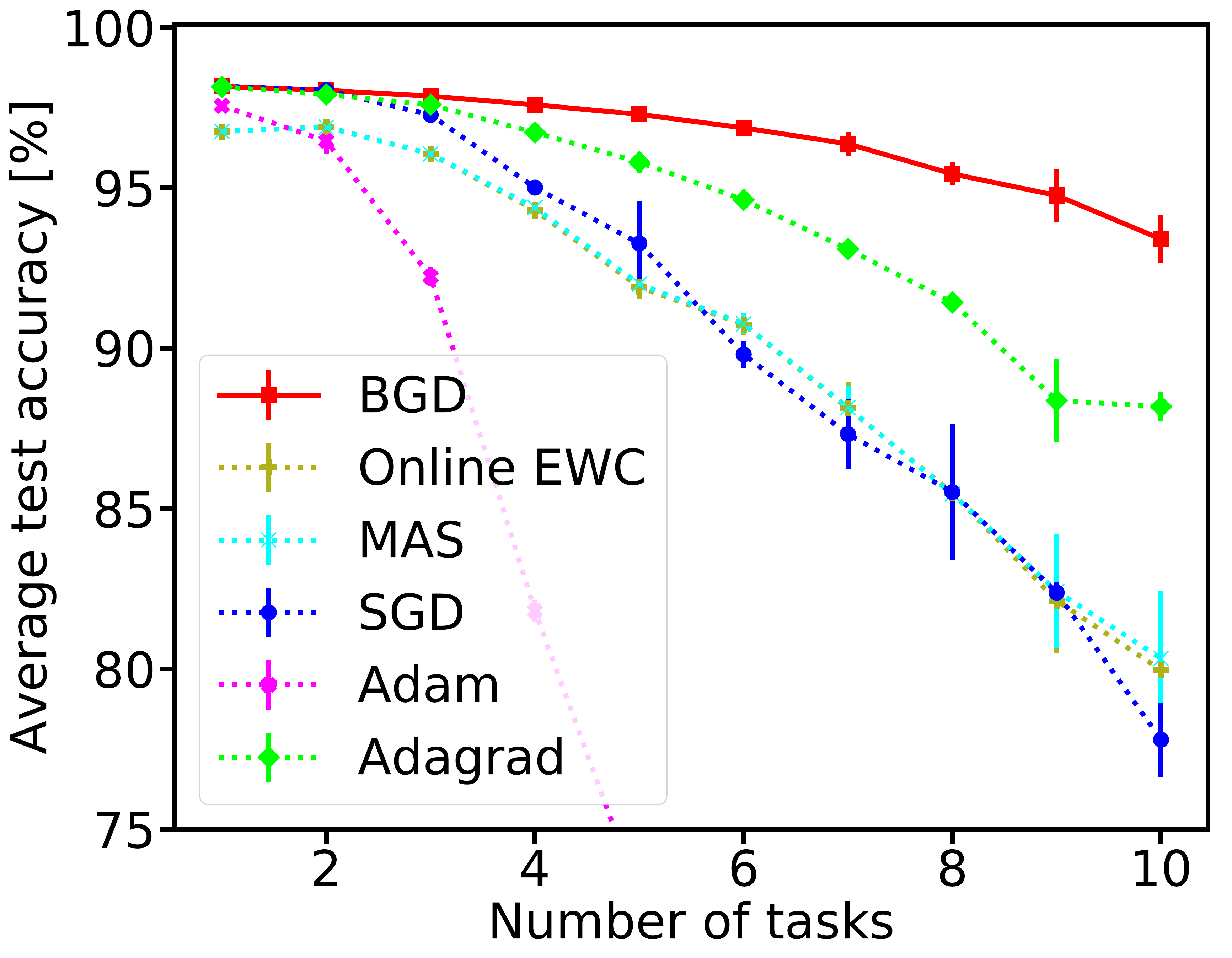}}
\caption{Results on Continuous Permuted MNIST. The scenario is continuous task agnostic continual learning: tasks are changing slowly over time as showed in Figure \ref{fig:tasks_distribution}. Reported accuracy is the average over three different runs, with error bars for STD.}
\label{fig:continuous_permuted_mnist}
\vspace{-1em}
\end{center}
\end{figure}

First, we present results on the continuous task-agnostic scenario, where the transition between tasks is done slowly over time (as in Figure \ref{fig:tasks_distribution}), so the task boundaries are undefined.
The output heads are shared among all tasks, task duration is $9380\times T$ iterations (corresponds to 20 epochs per task), and the algorithms are not given with the number of tasks.
Results of this scenario on permuted MNIST for different numbers of tasks presented in Figure \ref{fig:continuous_permuted_mnist}.
Reported accuracy is the average test accuracy over all tasks. BGD is able to maintain high average accuracy.

\paragraph{Inapplicability of other algorithms}
In task-agnostic scenarios, previous methods of continual learning are generally inapplicable, as they rely on taking some action on task switch.
The task switch cannot be easily guessed because we assume no knowledge on the number of tasks.
Nevertheless, one trivial adaptation is to take the core action every iteration instead of every task switch. 
Doing so is impractical for most algorithms due to computational complexity, but we have succeeded to run both Online EWC and MAS with such an adaptation.

\subsection{Discrete task agnostic scenario}

We use discrete task-agnostic scenario on Permuted MNIST to support our hypothesis of how BGD works in continual learning (subsection \ref{theoretical_properties}).

\begin{figure}[h]
\begin{center}
\includegraphics[width=0.8\columnwidth]{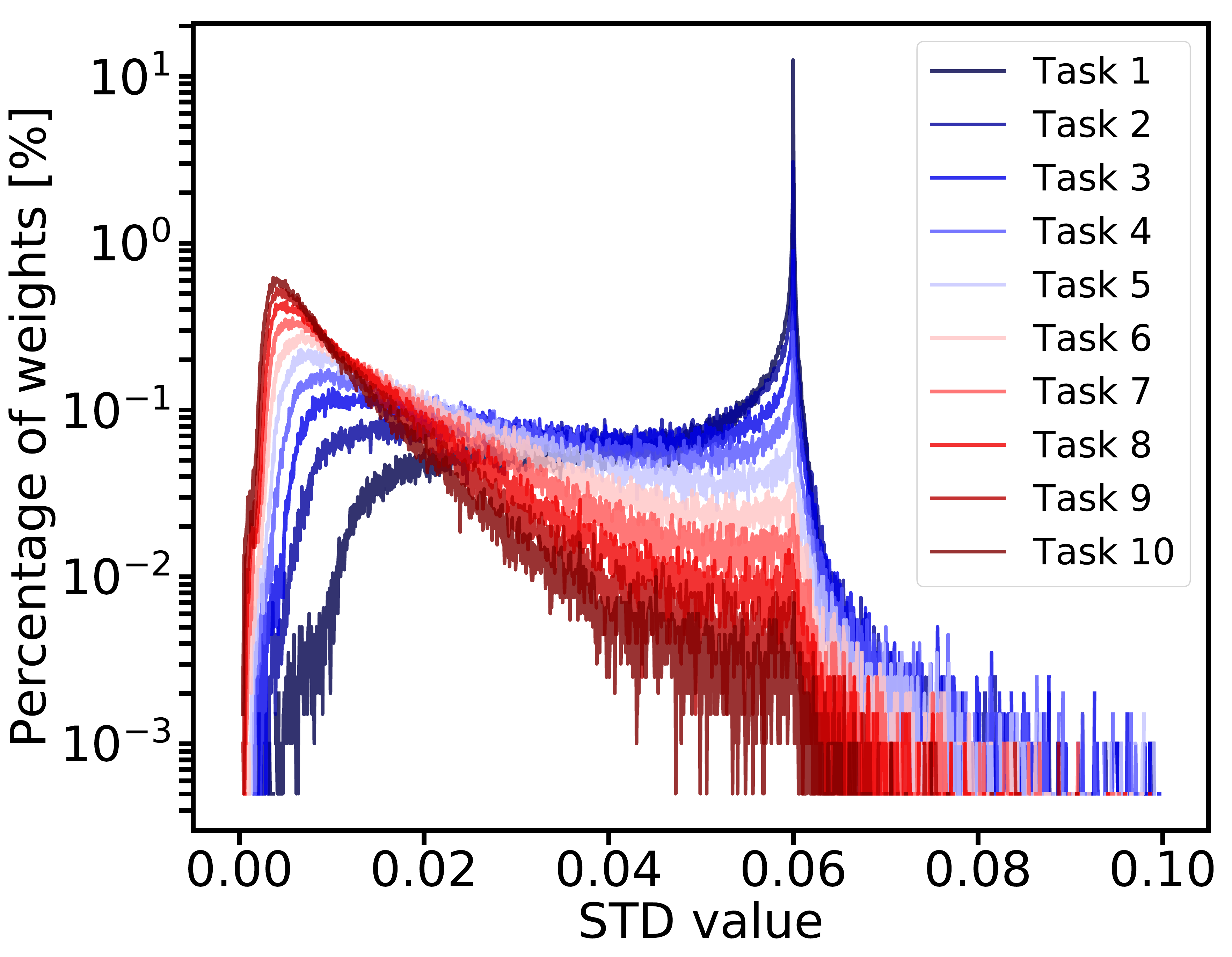}
\caption{The histogram of STD values at the end of the training process of each task, and the initial STD value is 0.06.}
\label{permuted_mnist_sigma}
\vspace{-1em}
\end{center}
\end{figure}

Figure \ref{permuted_mnist_sigma} shows the histogram of STD values at the end of the training process of each task.\footnote{BGD is unaware to the task identity, but we as the programmers know when a task is switched and recorded the STD values on that point.}
The results show that after the first task, a large portion of the weights have STD value which is close to the initial value of 0.06, while a small fraction of them have a much lower value. As training progresses,
more weights are assigned with STD values much lower than the initial value, but higher by at least a factor of 10 then the minimal value of $\sigma$, which is $\sim 10^{-4}$.
These results support our hypothesis in subsection \ref{theoretical_properties} that only a small part of the weights is essential for each task, and as training progresses the percentage of weights with small STD increases as more tasks are seen. It also shows that we do not collapse to point estimation. 

\begin{figure}[h]
\begin{center}
\includegraphics[width=0.8\columnwidth]{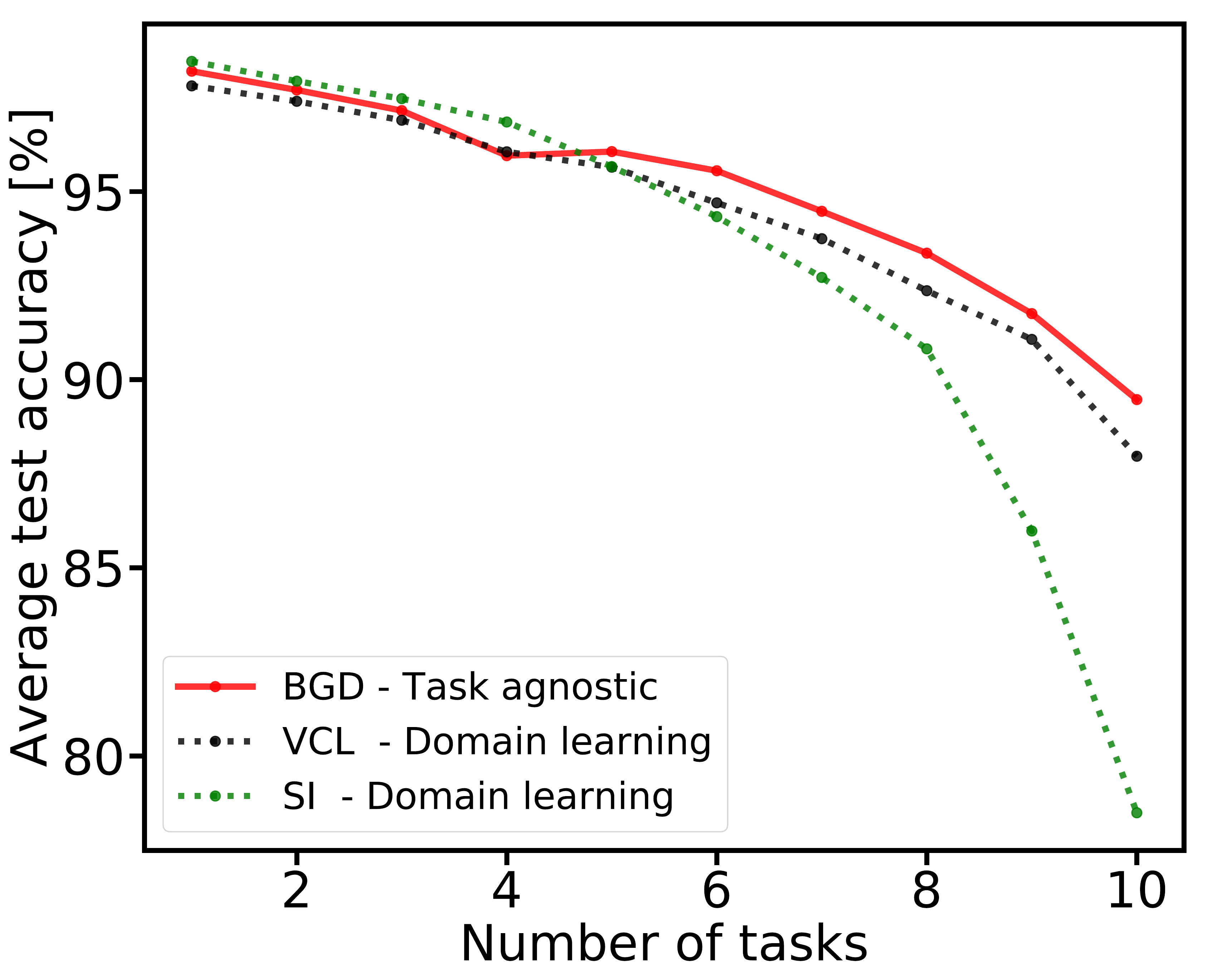}
\caption{The average test accuracy on permuted MNIST vs. the number of tasks. BGD (red), VCL (black) and SI (green). We used mini-batch of size 128 and 300 epochs for all the algorithms. Note that, in contrast to VCL and SI, BGD is task-agnostic (i.e. unaware of tasks changing), while still significantly alleviates catastrophic forgetting.}
\label{permuted_mnist_300}
\vspace{-1em}
\end{center}
\end{figure}

In terms of performance, BGD is compared with two diagonal methods: SI and VCL.\footnote{In \cite{zenke2017continual} the author compared SI and EWC on permuted MNIST and showed similar performance.}
Average accuracy on seen tasks is reported in Figure \ref{permuted_mnist_300}.
SI and VCL are evaluated in a domain learning scenario, since they use the knowledge on task switch during training.
BGD is evaluated in a discrete task-agnostic learning scenario, which means it has no information on task schedule.
Despite being evaluated on a more challenging scenario, BGD performs on-par with SI and VCL.
We find this highly encouraging, as naively one would expect task-agnostic algorithms to perform significantly worse, as they have less (important) information.

Another observation from this experiment is that training duration affects the performance of the algorithms.
The results reported in Figure \ref{permuted_mnist_300} are with 300 epochs per task. When training for a shorter duration, 20 epochs per task, SI achieves better performance than others (but, they all remain on-par), see Appendix \ref{appendix:Discrete permuted MNIST} for 20 epochs experiment results.

\subsection{``Labels trick'' for class learning}
\label{subsection:labels_trick_experiments}
In this experiment, we show that by applying the ``labels trick'' presented in this paper, we improve baseline results of ``class learning''. Task identity is known during training (using the labels), but not during testing, and the tasks do not share the output head ($T \times C$ heads, or more specifically $5 \times 2$ heads). We followed~\cite{hsu2018re} setup, and trained each task for four epochs (relatively short duration). Results on Split MNIST are reported in Table \ref{tbl:split_mnist_exp_labels_trick}. Labels trick improves results of all tested algorithms in such scenario.

\begin{table}[h]
\centering
\caption{Results for ``labels trick'' on Split MNIST. The average accuracy (\%) on all tasks at the end of overall training. Each value is the average of 10 runs. Class learning results are as reported in~\cite{hsu2018re}.}
\label{tbl:split_mnist_exp_labels_trick}
\begin{tabular}{cccc}
\toprule
Mehod                        & Class learning & + Labels trick \\ \midrule
 Adam                    & 19.71 $\pm$ 0.08  & 37.52 $\pm$ 3.1         \\
 SGD                     & 19.46 $\pm$ 0.04  & 52.71 $\pm$ 2.92           \\
 Adagrad                 & 19.82 $\pm$ 0.09  & 50.85 $\pm$ 4.69           \\ 
 L2                      & 22.52 $\pm$ 1.08  & 52.49 $\pm$ 3.14           \\ 
\midrule
 EWC                     & 19.80 $\pm$ 0.05  & 37.98 $\pm$ 2.24              \\ 
 Online EWC              & 19.77 $\pm$ 0.04  & 37.7  $\pm$ 3.27               \\ 
 SI                      & 19.67 $\pm$ 0.09  & 58.44 $\pm$ 3.04           \\
 MAS                     & 19.52 $\pm$ 0.29  & 60.43 $\pm$ 3.29           \\
 BGD (this paper)        & 19.64 $\pm$ 0.03  & 46.34 $\pm$ 2.36           \\ 
\midrule
Offline (upper bound)   & \multicolumn{2}{c}{97.53 $\pm$ 0.30}          \\
\bottomrule
\end{tabular}
\end{table}

\subsection{Additional experiments}
We provide additional evaluation of BGD using various continual learning experiments conducted in other works.
Those experiments are all done in one of the three basic scenarios, and some use more complex datasets such as CIFAR10/100. The algorithms which BGD is compared with are using the information on task-switch. Nevertheless, BGD results are on par in most cases. Those experiments can be found in Appendix \ref{appendix:Additional experiments}, and includes:
\vspace{-0.29cm}
\begin{enumerate}
    \setlength{\parskip}{0pt} \setlength{\itemsep}{0pt plus 2pt}
    \item Task, domain and class learning on Split MNIST and Permuted MNIST as in \cite{hsu2018re}.
    \item Task learning on ``vision datasets'' mixture (MNIST, notMNIST, FashionMNIST, SVHN and CIFAR10), as in~\cite{ritter2018online}.
    \item Task learning on CIFAR10 and subsets of CIFAR100, as in~\cite{zenke2017continual}.
\end{enumerate}
\vspace{-0.29cm}
In addition, as a sanity check to BGD, in Appendix \ref{appendixClassification} we show classification (single task) results on MNIST and on CIFAR10, using networks with Batch-Normalization \cite{ioffe2015batch}.
BGD can be easily applied to such networks. Lastly, in Appendix \ref{appendix:5000 epochs training}, we show that BGD does not experience underfitting and over-pruning, as shown by  \cite{trippe2018overpruning} for BBB. We show that by 5000 epochs training on MNIST and CIFAR10.

\section{Discussion and future work}
In this work, we address a new scenario of continual learning --- task-agnostic continual learning. We show that our algorithm is able to mitigate the notorious catastrophic forgetting phenomena, without being aware of task-switch. This important property can allow future models to better adapt to new tasks without explicitly instructed to do so, enabling them to learn in realistic continual learning settings. 

Our method, Bayesian Gradient Descent (BGD), show competitive results with various state-of-the-art continual learning algorithms on basic scenarios, while being able to work also in task-agnostic scenarios. It relies on solid theoretical foundations, and can be applied easily to advanced DNN architectures, including convolution layers and Batch-Normalization, and works on a broad variety of datasets.

Besides being a continual learning method, BGD had better classification accuracy than previous Bayesian methods on MNIST, and was on par with SGD on CIFAR10 (without Batch-Normalization).
Despite being an online version of the variational Bayes approach of \cite{blundell2015weight}, it does not seem to have the underfitting and over-pruning issues previously observed in the variational Bayes approach of \cite{blundell2015weight}. This allowed us to scale this approach beyond MNIST, to CIFAR10. The BGD approach is also closely related to the assumed density filtering approach of \cite{soudry2014expectation,hernandez2015probabilistic}. However, these methods rely on certain analytic approximations which are not easily applicable to different neural architectures (e.g. convnets).  In contrast,  it is straightforward to implement BGD for any neural architecture. This approach is also somewhat similar to the Stochastic Gradient Langevin Dynamics (SGLD) approach \citep{welling2011bayesian,balan2015bayesian}, in the sense that we use multiple copies of the network during training. However, in contrast to the SGLD approach, we are not required to store all copies of the networks for inference, but only two parameters for each weight ($\mu$ and $\sigma$).

There are many possible extensions and uses of BGD, which were not explored in this work. In this work, our variational approximation used a diagonal Gaussian distribution. This assumption may be relaxed in the future to non-diagonal or non-Gaussian (e.g., mixtures) distributions, to allow better flexibility during learning, however, such an extension is not trivial.
Lastly, the ``labels trick'' presented in this paper, places foundations for future works to improve performance in a class learning scenario. By proposing a method for accurate task prediction during testing, one can completely close the performance gap between class and task learning.

\subsubsection*{Acknowledgments}

The authors are grateful to R. Amit, M. Shpigel Nacson, N. Merlis, O. Linial, B. Shuval, and O. Rabinovich for helpful comments on the manuscript. This research was supported by the Israel Science foundation (grant No. 31/1031), and by the Taub foundation. A Titan Xp used for this research was donated by the NVIDIA Corporation.

\bibliographystyle{icml2019}

\clearpage
\appendix

\section{Derivation of \eqref{eq:mu implicit solution} and \eqref{eq:sigma implicit solution}}
\label{appendix:Derivation of eqs}
In this section we provide additional details on the derivation of \eqref{eq:mu implicit solution} and \eqref{eq:sigma implicit solution}. The objective function is $f\left(\bgdtheta,\phi\right)=\log\left(q_{n}\left(\bgdtheta|\phi\right)\right)-\log\left(q_{n-1}\left(\bgdtheta\right)\right)+\mathrm{L}_n\left(\bgdtheta\right)$,
where:
\begin{align}
\log\left(q_{n}\left(\bgdtheta|\phi\right)\right) & =-\frac{N}{2}\log\left(2\pi\right)-\sum_{k}\log\left(\sigma_{k}\right) \nonumber \\ &\quad -\sum_{k}\frac{1}{2\sigma_{k}^{2}}\left(\theta_{k}-\mu_{k}\right)^{2},
\end{align}
\begin{align}
\log\left(q_{n-1}\left(\bgdtheta\right)\right) & =-\frac{N}{2}\log\left(2\pi\right)-\sum_{k}\log\left(v_{k}\right)\nonumber \\ &\quad-\sum_{k}\frac{1}{2v_{k}^{2}}\left(\theta_{k}-m_{k}\right)^{2}.
\end{align}
We derive \eqref{eq:mu implicit solution} by using the first-order necessary conditions for the optimal $\mu_{i}$:
\begin{equation}
\mathbb{E}_{\varepsilon}\left[\frac{\partial f\left(\bgdtheta,\phi\right)}{\partial\theta_{i}}\frac{\partial\theta_{i}}{\partial\mu_{i}}+\frac{\partial f\left(\bgdtheta,\phi\right)}{\partial\mu_{i}}\right]=0\,.
\end{equation}
Substituting the derivatives, we obtain:
\begin{align}
&\mathbb{E}_{\varepsilon}\bigg[-\frac{1}{\sigma_{i}^{2}}\left(\theta_{i}-\mu_{i}\right)+\frac{1}{v_{i}^{2}}\left(\theta_{i}-m_{i}\right)\nonumber \\
&\qquad\, \!\!\! +\frac{\partial \mathrm{L}_n\left(\bgdtheta\right)}{\partial\theta_{i}}+\frac{1}{\sigma_{i}^{2}}\left(\theta_{i}-\mu_{i}\right)\bigg] \nonumber \\
&\quad= \frac{1}{v_{i}^{2}}\left(\mu_{i}-m_{i}\right)+E_{\varepsilon}\left[\frac{\partial \mathrm{L}_n\left(\bgdtheta\right)}{\partial\theta_{i}}\right]  =0\,.
\end{align}
And so we obtained \eqref{eq:mu implicit solution}.

Next, we derive \eqref{eq:sigma implicit solution}, using the first-order necessary conditions for optimal $\sigma_{i}$:
\begin{equation}
\mathbb{E}_{\varepsilon}\left[\frac{\partial f\left(\bgdtheta,\phi\right)}{\partial\theta_{i}}\frac{\partial\theta_{i}}{\partial\sigma_{i}}+\frac{\partial f\left(\bgdtheta,\phi\right)}{\partial\sigma_{i}}\right]=0\,.\label{eq:18}
\end{equation}
Substituting the derivatives we obtain:
\begin{align}
&\mathbb{E}_{\varepsilon}\bigg[\left(-\frac{1}{\sigma_{i}^{2}}\left(\theta_{i}-\mu_{i}\right)+\frac{1}{v_{i}^{2}}\left(\theta_{i}-m_{i}\right)+\frac{\partial \mathrm{L}_n\left(\bgdtheta\right)}{\partial\theta_{i}}\right)\varepsilon_{i}\nonumber \\
&\qquad -\frac{1}{\sigma_{i}}+\frac{1}{\sigma_{i}^{3}}\left(\theta_{i}-\mu_{i}\right)^{2}\bigg] \nonumber \\
&\quad =-\frac{1}{\sigma_{i}}+\frac{\sigma_{i}}{v_{i}^{2}}+\mathbb{E}_{\varepsilon}\left[\frac{\partial \mathrm{L}_n\left(\bgdtheta\right)}{\partial\theta_{i}}\varepsilon_{i}\right]  =0\,.\label{eq:19}
\end{align}
We get a quadratic equation for $\sigma_{i}$
\begin{equation}
\sigma_{i}^{2}+\sigma_{i}v_{i}^{2}E_{q\left(\varepsilon\right)}\left[\frac{\partial \mathrm{L}_n\left(\bgdtheta\right)}{\partial\theta_{i}}\varepsilon_{i}\right]-v_{i}^{2}=0\,.
\end{equation}
Since $\sigma_{i}>0$, the solution is \eqref{eq:sigma implicit solution}.

\section{Proof of Theorem \ref{theorem}}
\label{appendix:Proof of Theorem}
\begin{proof}
We define $\theta_{j}=\mu_{j}+\varepsilon_{j}\sigma_{j}$ where  $\varepsilon_{j}\sim\mathcal{N}\left(0,1\right)$.
According to the smoothing theorem, the following holds
\begin{equation}
\mathbf{\mathbb{E}}_{\varepsilon}\left[\frac{\partial \mathrm{L}_n\left(\bgdtheta\right)}{\partial\theta_{i}}\varepsilon_{i}\right]=\mathbb{E}_{\varepsilon_{j\neq i}}\left[\mathbf{\mathbb{E}}_{\varepsilon_{i}}\left[\left.\frac{\partial \mathrm{L}_n\left(\bgdtheta\right)}{\partial\theta_{i}}\varepsilon_{i}\right|\varepsilon_{j\neq i}\right]\right]\,.
\end{equation}
The conditional expectation is:
\begin{equation}
\mathbf{\mathbb{E}}_{\varepsilon_{i}}\left[\left.\frac{\partial \mathrm{L}_n\left(\bgdtheta\right)}{\partial\theta_{i}}\varepsilon_{i}\right|\varepsilon_{j\neq i}\right]=\intop_{-\infty}^{\infty}\frac{\partial \mathrm{L}_n\left(\bgdtheta\right)}{\partial\theta_{i}}\varepsilon_{i}f_{\varepsilon_{i}}\left(\varepsilon_{i}\right)d\varepsilon_{i}\,,
\end{equation}
where $f_{\varepsilon_{i}}$ is the probability density function of a standard normal distribution. Since $f_{\varepsilon_{i}}$ is an even function
\begin{align}
& \mathbf{\mathrm{E}}_{\varepsilon_{i}}\left[\left.\frac{\partial \mathrm{L}_n\left(\bgdtheta\right)}{\partial\theta_{i}}\varepsilon_{i}\right|\varepsilon_{j\neq i}\right] \nonumber  \\
 &\quad =  \intop_{0}^{\infty}\frac{\partial \mathrm{L}_n\left(\mu_{i}+\varepsilon_{i}\sigma_{i},\theta_{-i}\right)}{\partial\theta_{i}}\varepsilon_{i}f_{\varepsilon_{i}}\left(\varepsilon_{i}\right)d\varepsilon_{i} \nonumber \\
& \qquad -\intop_{0}^{\infty}\frac{\partial \mathrm{L}_n\left(\mu_{i}-\varepsilon_{i}\sigma_{i},\theta_{-i}\right)}{\partial\theta_{i}}\varepsilon_{i}f_{\varepsilon_{i}}\left(\varepsilon_{i}\right)d\varepsilon_{i}\,. \label{eq:expectation}
\end{align}

Now, since $\mathrm{L}_n\left(\bgdtheta\right)$ is strongly convex function
with parameter $m_n>0$ and continuously differentiable function over
$\mathbb{R}^{d}$, the following holds $\forall\bgdtheta_{1},\bgdtheta_{2}\in\mathbb{R}^{d}$:
\begin{equation}
\left(\nabla \mathrm{L}_n\left(\bgdtheta_{1}\right)-\nabla \mathrm{L}_n\left(\bgdtheta_{2}\right)\right)^{T}\left(\bgdtheta_{1}-\bgdtheta_{2}\right)\geq m_n\left\| \bgdtheta_{1}-\bgdtheta_{2}\right\|_2^2\,.
\end{equation}
For $\bgdtheta_{1},\bgdtheta_{2}$ such that
\begin{align}
\left(\bgdtheta_{1}\right)_{j}=\begin{cases}
\left(\bgdtheta_{2}\right)_{j}, & j\neq i\\
\mu_{i}+\varepsilon_{i}\sigma_{i}, & j=i\,,
\end{cases}
\end{align}
\begin{align}
\left(\bgdtheta_{2}\right)_{j}=\begin{cases}
\left(\bgdtheta_{1}\right)_{j}, & j\neq i\\
\mu_{i}-\varepsilon_{i}\sigma_{i}, & j=i\,,
\end{cases}
\end{align}
the following holds:
\begin{equation}
\left(\frac{\partial \mathrm{L}\left(\bgdtheta_{1}\right)}{\partial\theta_{i}}-\frac{\partial \mathrm{L}\left(\bgdtheta_{2}\right)}{\partial\theta_{i}}\right)\varepsilon_{i}\geq 2m_n\sigma_i\varepsilon_{i}^2\,.
\end{equation}
Therefore, substituting this inequality into \eqref{eq:expectation}, we obtain:
\begin{equation}
\mathbf{\mathbb{E}}_{\varepsilon}\left[\frac{\partial \mathrm{L}\left(\bgdtheta\right)}{\partial\theta_{i}}\varepsilon_{i}\right]\geq m_n\sigma_i>0\,.
\end{equation}
\end{proof}

\section{Datasets}
\label{appendix:Datasets}

\textbf{MNIST} is a database of handwritten digits, which has a training set of 60,000
examples, and a test set of 10,000 examples --- each a $28 \times 28$ image.
The image is labeled with a number in the range of 0 to 9.

\textbf{Permuted MNIST}
is a set of tasks constructed by a random
permutation of MNIST pixels. Each task has a different permutation of
pixels from the previous one. 

\textbf{Split MNIST}
is a set of tasks constructed by taking pairs of digits from MNIST dataset. For example: the first task is classifying between the digits 0 and 1, the second task is classifying between the digits 2 and 3 etc. The total number of tasks in this experiment is five.

\textbf{CIFAR10} 
is a dataset which consists of 60000 $32 \times 32$ color images in 10 classes, with 6000 images per class. There are 50000 training images and 10000 test images. 

\textbf{CIFAR100} 
is a dataset which consists of 60000 $32 \times 32$ color images in 100 classes, with 600 images per class \citep{krizhevsky2009learning}. There are 50000 training images and 10000 test images. 
The 100 classes in the CIFAR-100 are grouped into 20 superclasses. Each image comes with a ``fine'' label (the class to which it belongs) and a ``coarse'' label (the superclass to which it belongs). 

\textbf{FashionMNIST} is a dataset comprising of $28 \times 28$ grayscale images of 70,000 fashion products from 10 categories. The training set has 60,000 images and the test set has 10,000 images.

\textbf{notMNIST} is a dataset with 10 classes, with letters A-J taken from different fonts. It has similar characteristics like MNIST - same pixel size ($28 \times 28$), grayscale images and comprised of 10-class.

\textbf{SVHN} is a real-world image dataset with 10 classes, 1 for each digit. It has RGB images of size $32 \times 32$ obtained from house numbers in Google Street View images.

\section{Implementation details}
\label{appendix:Implementation details}

We initialize the mean of the weights $\mu$ by sampling from a Gaussian
distribution with a zero mean and a variance of $2/(n_{\mathrm{input}}+n_{\mathrm{output}})$ as in \cite{glorot2010understanding}, unless stated otherwise.
We use $10$ Monte Carlo samples to estimate the expected gradient during training, and average the accuracy of 10 sampled networks during testing, unless stated otherwise.

\begin{figure}[ht]
\begin{center}
\centerline{\includegraphics[width= 0.8\columnwidth]{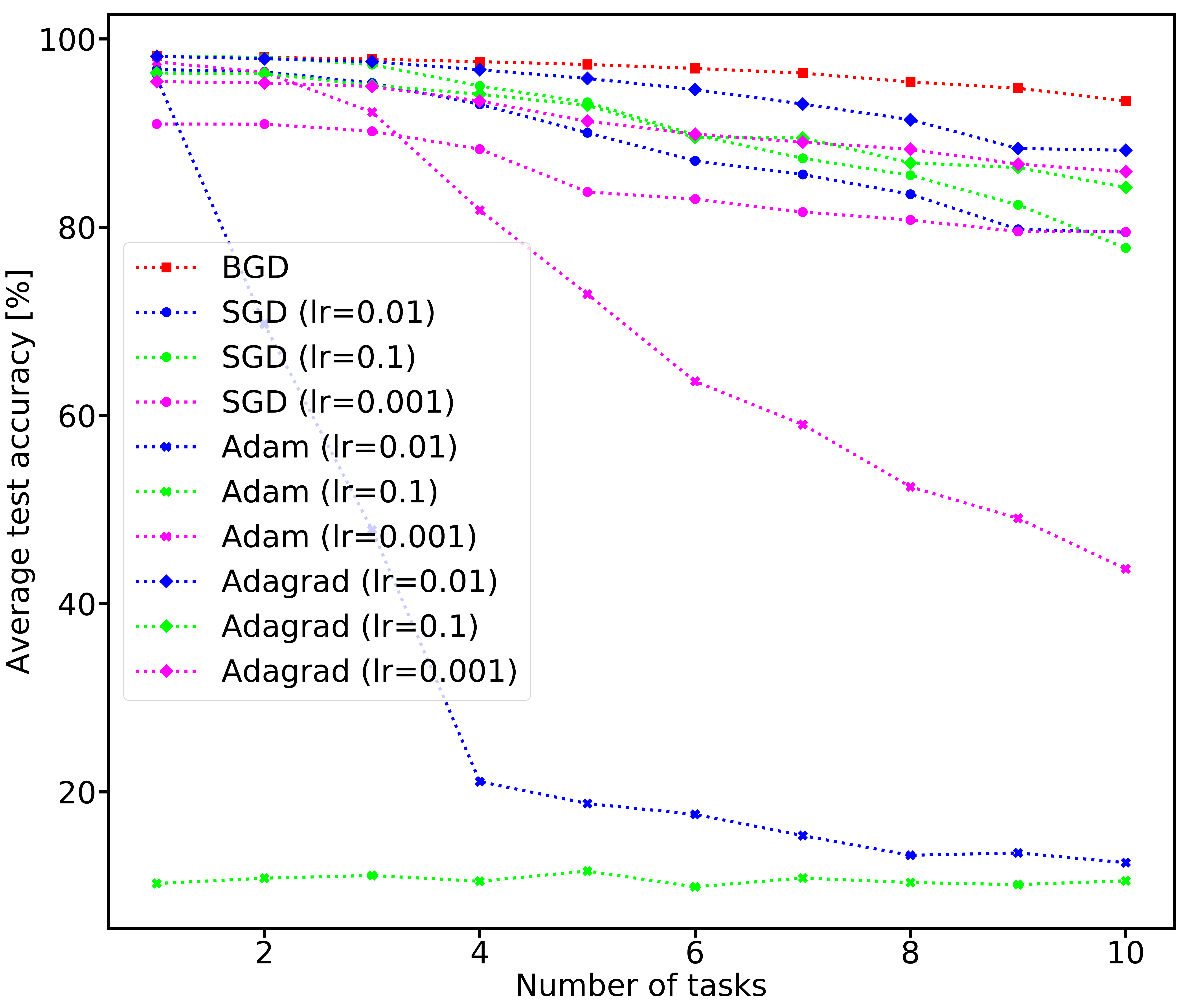}}
\caption{Hyper-parameters search for continuous permuted MNIST experiment.}
\label{cont_permuted_hyperparameter_search}
\vspace{-1em}
\end{center}
\end{figure}

\paragraph{Continuous permuted MNIST}
We use a fully connected network with 2 hidden layers of width 200.
Following \cite{hsu2018re}, the original MNIST images are padded with zeros to match size of $32 \times 32$. The batch size is 128, and we sample with replacement due to the properties of the continuous scenario (no definition for epoch as task boundaries are undefined).
We conducted a hyper-parameter search for all algorithms and present the best, see 
Figure \ref{cont_permuted_hyperparameter_search}.
Results are averaged over three different seeds (2019, 2020, and 2021).
STD init for BGD is $0.06$ and $\eta = 1$.
For Online EWC and MAS we used the following combinations of hyper-parameters: LR of 0.01 and 0.0001, regularization coefficient of 10, 0.1, 0.001, 0.0001, and optimizer SGD and Adam. The best results for online EWC were achieved using LR of 0.01, regularization coefficient of 0.1 and SGD optimizer.
For MAS, best results achieved using LR of 0.01, regularization coefficient of 0.001 and SGD optimizer.

\begin{figure}[h]
\begin{center}
\centerline{\includegraphics[width= 0.8\columnwidth]{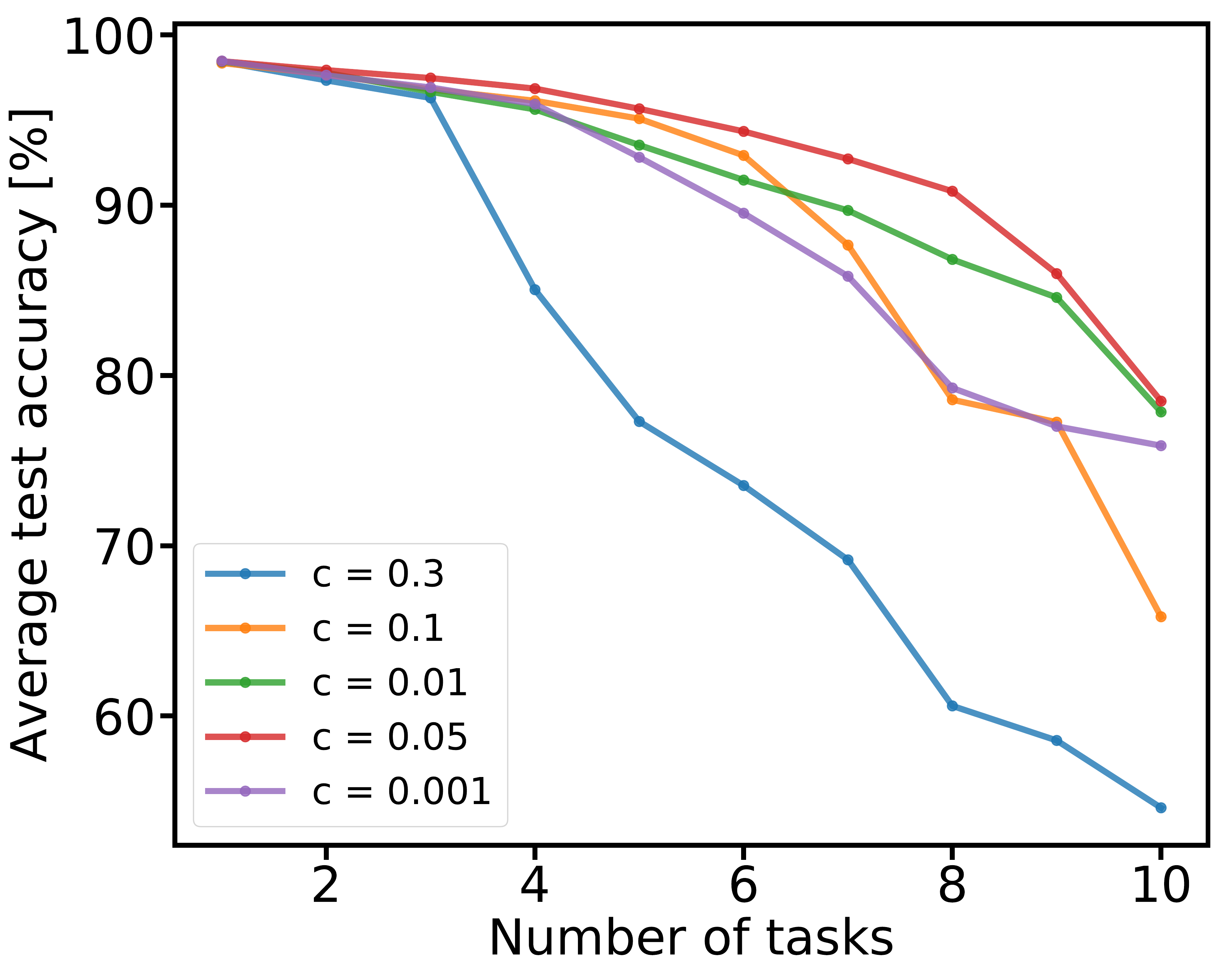}}
\caption{SI average test accuracy on permuted MNIST vs. the number of tasks with different hyper-parameter values.}
\label{si hyper-parameter values}
\vspace{-1em}
\end{center}
\end{figure}

\paragraph{Discrete permuted MNIST}
We use a fully connected neural network
with 2 hidden layers of 200 width, ReLUs as activation functions
and softmax output layer with 10 units. We trained the network using
Bayesian Gradient Descent. The preprocessing is the same as with MNIST classification, but we used a training set of 60,000 examples.
BGD was trained with mini-batch of size 128 and $\eta=1$ for 300 epochs. 
We run an hyper-parameters tuning for SI using $c=(0.3,0.1,0.05,0.01,0.001)$ and select the best one (c = 0.05) as a baseline (See Figure \ref{si hyper-parameter values}).

\paragraph{Labels trick for class learning on Split MNIST}
We followed \cite{hsu2018re} setup. The network is a fully connected network with 2 hidden layers of width 400. The batch size is 128, and each task is trained for 4 epochs. The original MNIST images are padded with zeros. STD init for BGD is $0.017$ and $\eta = 1$. Results are averaged over ten different seeds (2019-2028).

\paragraph{CIFAR10/CIFAR100 continual learning}
Batch size is 256. On BGD we used MAP for inference method, initial STD was 0.019 and $\eta=2$. For SGD we used constant learning rate of 0.01 and for SI we used $c=0.01$. Hyper parameters for all algorithms were chosen after using grid search and taking the optimal value.

\paragraph{Vision datasets mix}
We followed the experiment as described in \cite{ritter2018online}. We use a batch size of 64, and we normalize the datasets to have zero mean and unit variance. The network architecture is LeNet like with 2 convolution layers with 20 and 50 channels and kernel size of 5, each convolution layer is followed by a Relu activations function and max pool, and the two layers are followed with fully connected layer of size 500 before the last layer.

SGD baseline was trained with a constant learning rate of 0.001 and ADAM used $\epsilon=10^{-8}$, LR of 0.001 and $(\beta_1,\beta_2)=(0.9, 0.999)$. BGD trained with initial STD of 0.02, $\eta$ set to 1 and batch size 64.

\paragraph{MNIST classification}
We use a fully
connected neural network with two hidden layers of various widths, ReLU's
as activation functions and softmax output layer with 10 units. We
preprocessed the data to have pixel values in the range of $\left[0,1\right]$.
We use mini-batch of size 128 and $\eta=1$. In addition, we used a validation set to find
the initialization value for $\sigma$. In order to compare the results of Bayesian Gradient Descent (BGD) with the results of Bayes By Backprop
%\footnote{We show results with the same prior as Bayesian Gradient Descent use.}
(BBB) algorithm by \cite{blundell2015weight}, we used a training set of 50,000 examples and validation set of 10,000 examples and present results of BBB with the same prior as BGD used. We train the network for 600 epochs. 

\paragraph{CIFAR10 classification}
SGD was trained with learning rate starting at 0.01 and divided by 10 every 100 epochs,  momentum was set to 0.9.
On both BGD and SGD we used data augmentation of random cropping and flipping. Both experiments use a batch size of 128 and 10 Monte Carlo iterations.
In the experiment with Batch-Normalization (BN) initial STD is 0.011, $\eta$ set to 8. $\mu$ is initialized using PyTorch 0.3.1 for convolution layers, bias is initialized to 0 and $\mu$ of BN layers scaling and shifting parameters are initialized to 1 and 0.
In the experiment without Batch-Normalization (BN) initial std is 0.015, $\eta$ set to 10. $\mu$ of convolution layers is initialized using He initialization.

\section{Complexity}
\label{appendix:complexity discussion}
BGD requires $\times 2$ more parameters compared to SGD, as it stores both the mean and the STD per weight.
In terms of time complexity, the major difference between SGD and BGD arises from the estimation of the expected gradients using Monte Carlo samples during training.
%the expected gradients estimation using Monte Carlo samples during training.
Since those Monte Carlo samples are completely independent the algorithm is embarrassingly parallel.

Specifically, given a mini-batch: for each Monte Carlo sample, BGD generates a random network using $\mu$ and $\sigma$, then making a forward-backward pass with the randomized weights.

Two main implementation methods are available (using 10 Monte Carlo samples as an example):
\begin{enumerate}
    \item
    Producing the (10) Monte Carlo samples sequentially, thus saving only a single randomized network in memory at a time (decreasing memory usage, increasing runtime).
    \item
    Producing the (10) Monte Carlo samples in parallel, thus saving (10) randomized networks in memory (increasing memory usage, decreasing runtime).
\end{enumerate}

We analyzed how the number of Monte Carlo iterations affects the runtime on CIFAR-10 using the first method of implementation (sequential MC samples). The results, reported in Table \ref{timinganalysis}, show that runtime is indeed a linear function of the number of MC iterations.
In the experiments we used a single GPU (GeForce GTX 1080 Ti).
\begin{table}
\caption{Average runtime of a single training epoch with different numbers of Monte Carlo samples. The MC iterations have linear effect on runtime. Less MC iterations does not affect accuracy. Accuracy reported in the table is on CIFAR10 classification task.}
\begin{center}
\begin{tabular}{c|c|c|c}
\textbf{MC iterations} & \textbf{Accuracy} & \textbf{Runtime}& \textbf{Vs. SGD} \\
 &  & [seconds] &
\tabularnewline
\hline 
SGD & N/A & 7.8 & $\times$1 \tabularnewline
\hline 
2 (BGD) & 89.31\% & 18.9 & $\times$ 2.4 \tabularnewline
\hline 
4 (BGD) & 89.32\% & 35.2 & $\times$ 4.5 \tabularnewline
\hline 
10 (BGD)& 89.04\% & 83.8 & $\times$ 10.7 \tabularnewline
\label{timinganalysis}
\end{tabular}
\end{center}
\end{table}

\section{Additional experiments}
\label{appendix:Additional experiments}
\subsection{Spilt MNIST and permuted MNIST}
\label{appendix:Spilt MNIST and permuted MNIST}
We followed~\cite{hsu2018re} setup, and compared BGD to several continual learning algorithms, on all three basic scenarios (task, domain, and class learning).
Results for Split MNIST reported on Table \ref{tbl:split_mnist_exp}, and for permuted MNIST on Table \ref{tbl:permuted_mnist_exp}.
Note that those experiments are with relatively short duration of training.

For Permuted MNIST experiments we used STD init of 0.02.
For task learning on Split MNIST we used STD init of 0.02, for domain learning we used 0.05, and 0.017 for class learning.

\subsection{Continual learning on vision datasets}
\label{appendix:vision dataset}
We followed \cite{ritter2018online} and challenged our algorithm with the vision datasets experiment. In this experiment, we train sequentially on MNIST, notMNIST, \footnote{Originally published at \\ http://yaroslavvb.blogspot.co.uk/2011/09/notmnist-dataset.html and downloaded from https://github.com/davidflanagan/notMNIST-to-MNIST} FashionMNIST, SVHN and CIFAR10 \citep{lecun1998gradient,xiao2017fashion,netzer2011reading,krizhevsky2009learning}.
Training is done in a sequential way with 20 epochs per task --- in epochs 1-20 we train on MNIST (first task), and on epochs 81-100 we train on CIFAR10 (last task).
All five datasets consist of about 50,000 training images from 10 different classes, but they differ from each other in various ways: black and white vs. RGB, letters and digits vs. vehicles and animals etc. See Appendix \ref{appendix:Datasets} for further details about the datasets.
We use the exact same setup as in \cite{ritter2018online} for the comparison --- LeNet-like \citep{lecun1998gradient} architecture with separated last layer for each task as in CIFAR10/CIFAR100 experiment. Results are reported on Table \ref{table:visiondatasets_results}.

\subsection{Continual learning on CIFAR10/CIFAR100}
\label{SPLIT CIFAR100}
We follow SI \citep{zenke2017continual} experiment on CIFAR10/CIFAR100 --- training sequentially on six tasks from CIFAR10 and CIFAR100.
The first task is the full CIFAR10 dataset, and the next five tasks are subsets of CIFAR100 with ten classes, where we train each task for 150 epochs.
As a reference, we present results of SGD and SI. We used the same network as in the original experiment, which consists of four convolutional layers followed by a fully connected layer with dropout shared by all six tasks.

\begin{figure}[ht]
\begin{center}
\includegraphics[width=0.8\columnwidth]{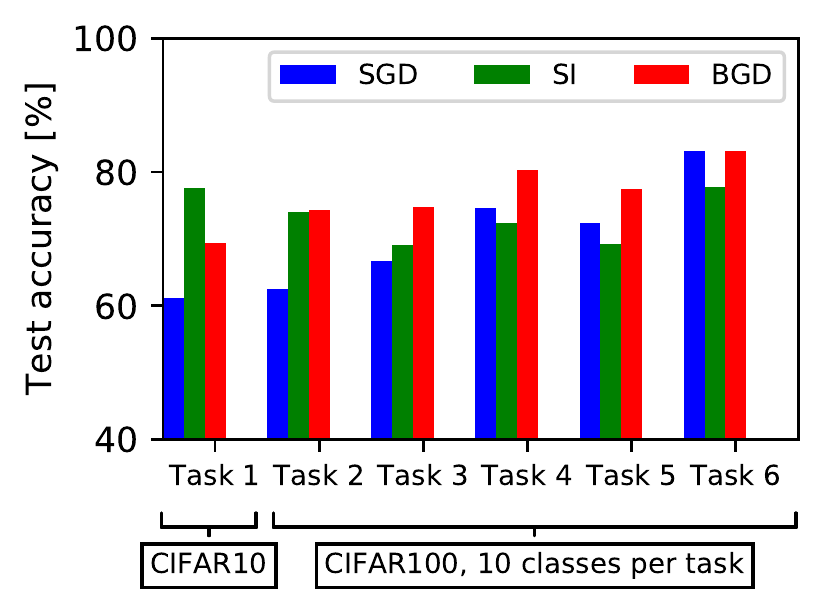}
\caption{Test accuracy per task on last epoch for continual learning on CIFAR10 and subsets of CIFAR100. Task 1 is the first seen task and task six is the last.}
\label{CIFAR100_catastrophic_forgetting}
\vspace{-1em}
\end{center}
\end{figure}

The results (Figure \ref{CIFAR100_catastrophic_forgetting}) show that the network is able attain a good balance between retaining reasonable accuracy on previous tasks and achieving high accuracy on newer ones. 

\begin{figure}[ht]
\begin{center}
\includegraphics[width=0.8\columnwidth]{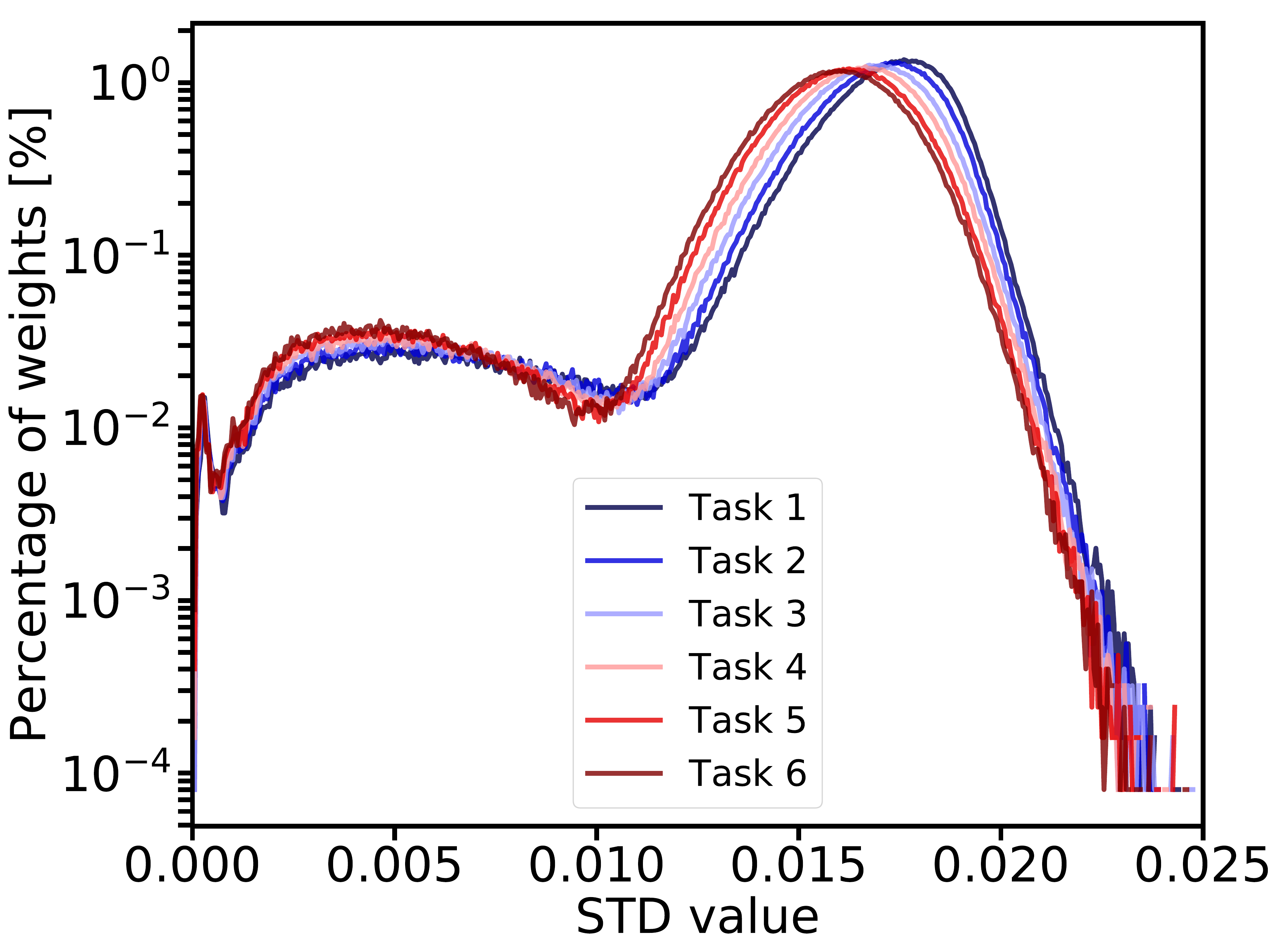}
\caption{The histogram of STD values at the end of the training process of each task, and the initial STD value is 0.019.}
\label{CIFAR100_catastrophic_forgetting_sigma}
\vspace{-1em}
\end{center}
\end{figure}

Figure \ref{CIFAR100_catastrophic_forgetting_sigma} shows the histogram of STD values at the end of the training process of each task. Similar to the results for permuted MNIST, as more tasks are seen the percentage of weights with STD values smaller than initial STD value (of 0.019) increases (the minimal value of $\sigma$ is $\sim 10^{-4}$).

\subsection{Discrete permuted MNIST}
\label{appendix:Discrete permuted MNIST}

\begin{figure}[h]
\begin{center}
\centerline{\includegraphics[width= 0.8\columnwidth]{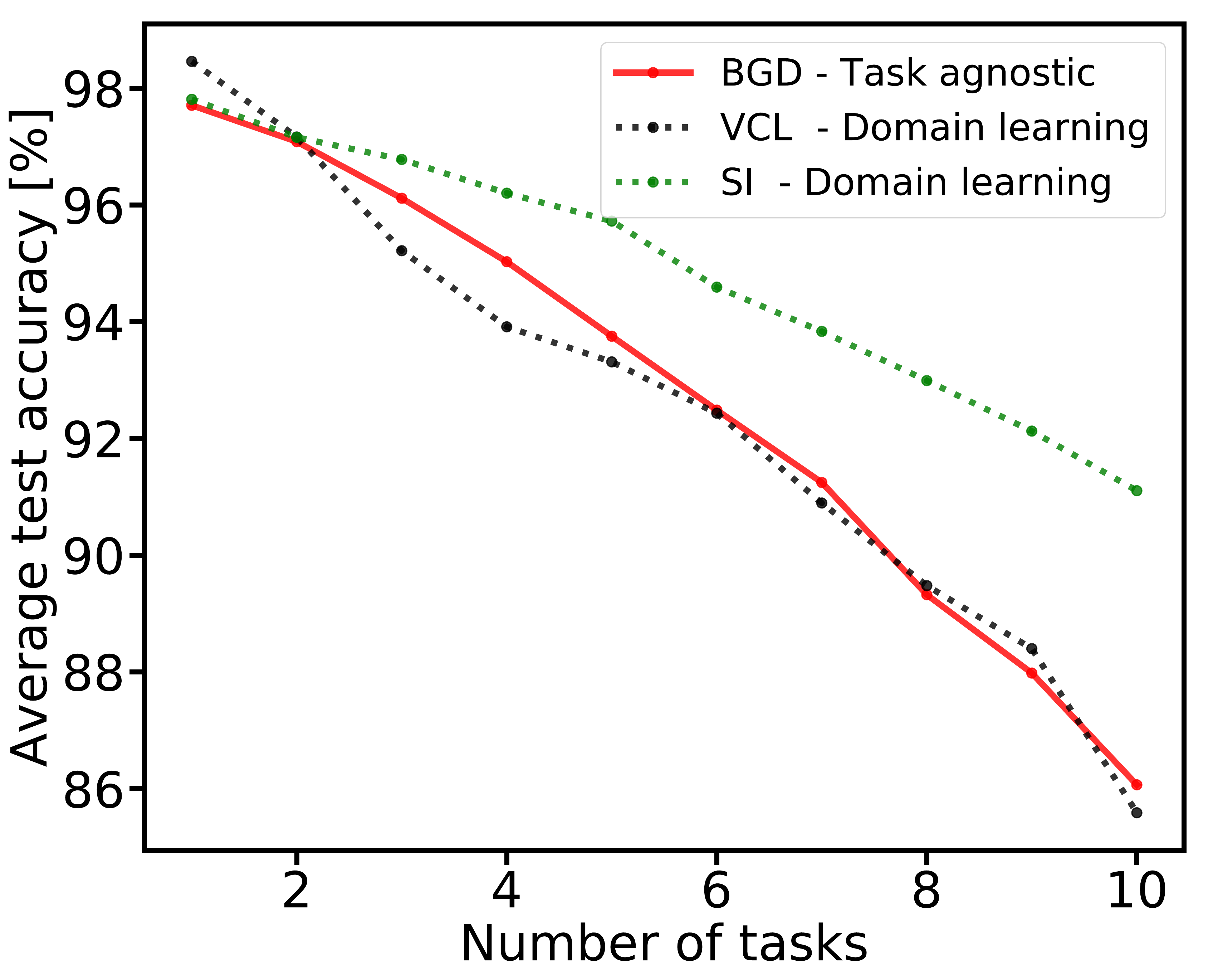}}
\caption{The average test accuracy on permuted MNIST vs. the number of tasks. BGD (red), VCL (black), SI (green) and SGD (blue), We used mini-batch of size 256 and 20 epochs for all the algorithms.}
\label{permuted_mnist_20_epochs}
\vspace{-1em}
\end{center}
\end{figure}

We ran an additional simulation with the same architecture on exactly the configuration as in \cite{zenke2017continual}, see Figure \ref{permuted_mnist_20_epochs}.

\subsection{Classification}
\label{appendixClassification}
As a sanity check, we compare BGD performance on single-task classification to recent Bayesian methods used for classification along with SGD and ADAM.

\paragraph{MNIST classification}
\begin{savenotes}
\begin{table}[!htbp]
\caption{Test accuracy for MNIST classification.}
\begin{center}
\begin{tabular}{c|c|c|c}
\textbf{Layer} & \textbf{BBB} & \textbf{BGD} & \textbf{SGD}\tabularnewline
\textbf{width} & \textbf{(Gaussian)} & &
\tabularnewline
\hline 
400 & 98.18\% & \textbf{98.26\%} & 98.17\% \tabularnewline
\hline 
800 & 98.01\% & \textbf{98.33\%} & 98.16\% \tabularnewline
\hline 
1200 & 97.96\% & \textbf{98.22\%} & 98.12\%
\label{mnist_acc}
\end{tabular}
\end{center}
\end{table}
\end{savenotes}

We compare BGD on MNIST with SGD and BBB~\citep{blundell2015weight}.
Table \ref{mnist_acc} summarizes test accuracy\footnote{Results of SGD and BBB in Table \ref{mnist_acc} for MNIST are as reported on~\cite{blundell2015weight}.} of networks with various widths. Overall, BGD performs better than BBB and SGD.

\paragraph{CIFAR10 classification}
We use CIFAR10 to evaluate BGD's performance on a more complex dataset and a larger network.
In this experiment, besides comparison to SGD and ADAM, we compare BGD to a group of recent advanced variational inference algorithms, including both diagonal and non-diagonal methods --- K-FAC~\citep{martens2015optimizing}, Noisy K-FAC and Noisy ADAM~\citep{zhang2017noisy}.

To do so, we followed the experiment described in~\cite{zhang2017noisy}. There, a VGG16-like\footnote{The detailed network architecture is 32-32-M-64-64-M-128-128-128-M-256-256-256-M-256-256-256-M-FC10, where each number represents the number of filters in a convolutional layer, and M denotes max-pooling - same as~\cite{zhang2017noisy}.}~\citep{simonyan2014very} architecture was used and data augmentation is applied. We present results both with Batch Normalization~\citep{ioffe2015batch} and without it in Table \ref{table:vgg16zhang_acc_and_ece}.

\begin{table}
\caption{Classification accuracy on CIFAR10 with modified VGG16 as in~\cite{zhang2017noisy}. All results except for BGD and ADAM are as reported in~\cite{zhang2017noisy}. [N/A] values are due to extreme instability.}
\begin{center}
\begin{tabular}{l|l|l}
    & \multicolumn{2}{c}{\textbf{Test Accuracy [\%]}}                   \\
\textbf{Method}          & \textbf{Without BN}                   & \multicolumn{1}{l}{\textbf{With BN}}           \\ \hline
\multicolumn{3}{l}{\rule{0pt}{3ex} \underline{Diagonal methods}} \\
BGD       & \textbf{88.41}                  & \multicolumn{1}{l}{91.07}    \\
SGD       & 88.35            & \multicolumn{1}{l}{\textbf{91.39}}    \\
ADAM       & 87.12           & \multicolumn{1}{l}{90.15}    \\
BBB \footnotemark       & 88.31            & \multicolumn{1}{l}{N/A} 
\\
NoisyADAM & 88.23            & \multicolumn{1}{l}{N/A}  \\
\multicolumn{3}{l}{\rule{0pt}{3ex} \underline{Non-diagonal methods}} \\
KFAC      & 88.89                  & \multicolumn{1}{l}{\textbf{92.13}}         \\
NoisyKFAC & \textbf{89.35}                  & \multicolumn{1}{l}{92.01} 
\label{table:vgg16zhang_acc_and_ece}
\end{tabular}
\end{center}
\end{table}
\footnotetext{A relaxed version of BBB was used, with $\lambda=0.1$ as described in~\cite{zhang2017noisy}.}

\subsection{5000 epochs training}
\label{appendix:5000 epochs training}
\paragraph{MNIST}
We use the MNIST classification experiment to demonstrate the convergence of the log-likelihood cost function and the histogram of STD values. We train a fully connected neural network with two hidden layers and layer width of 400 for 5000 epochs.

\begin{figure}[ht]
\begin{center}
\centerline{\includegraphics[width= 0.8\columnwidth]{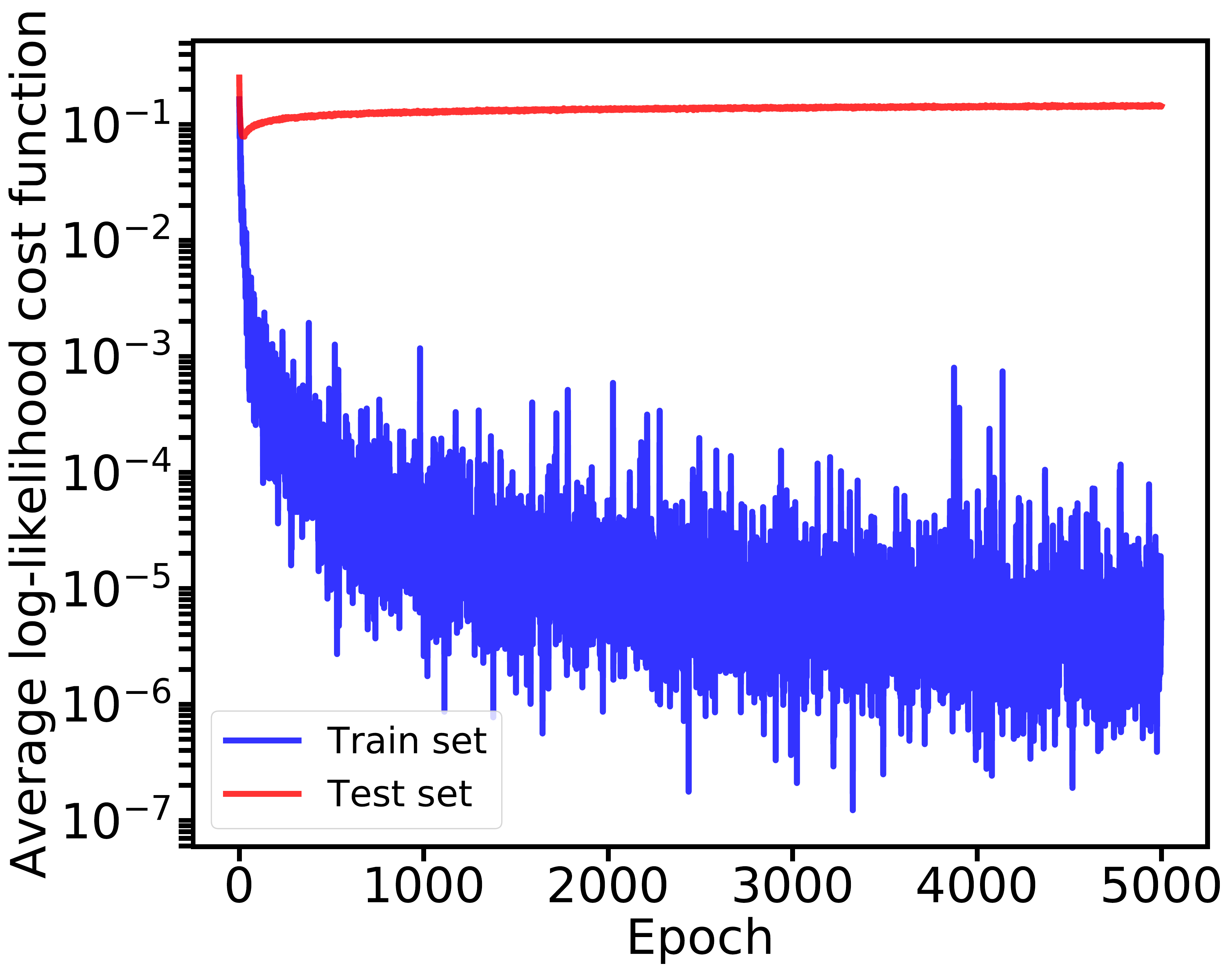}}
\caption{Average log-likelihood cost function of the train set and the test set - layer width 400.}
\label{loss_mnist}
\vspace{-1em}
\end{center}
\end{figure}

Figure \ref{loss_mnist} shows the log-likelihood cost function of the training set and the test set. As can be seen, the log-likelihood cost function on the training set decreases during the training process and converges to a low value. Thus, BGD does not experience underfitting and over-pruning as was shown by~\cite{trippe2018overpruning} for BBB.

\begin{figure}[!h]
\begin{center}
\centerline{\includegraphics[width= 0.8\columnwidth]{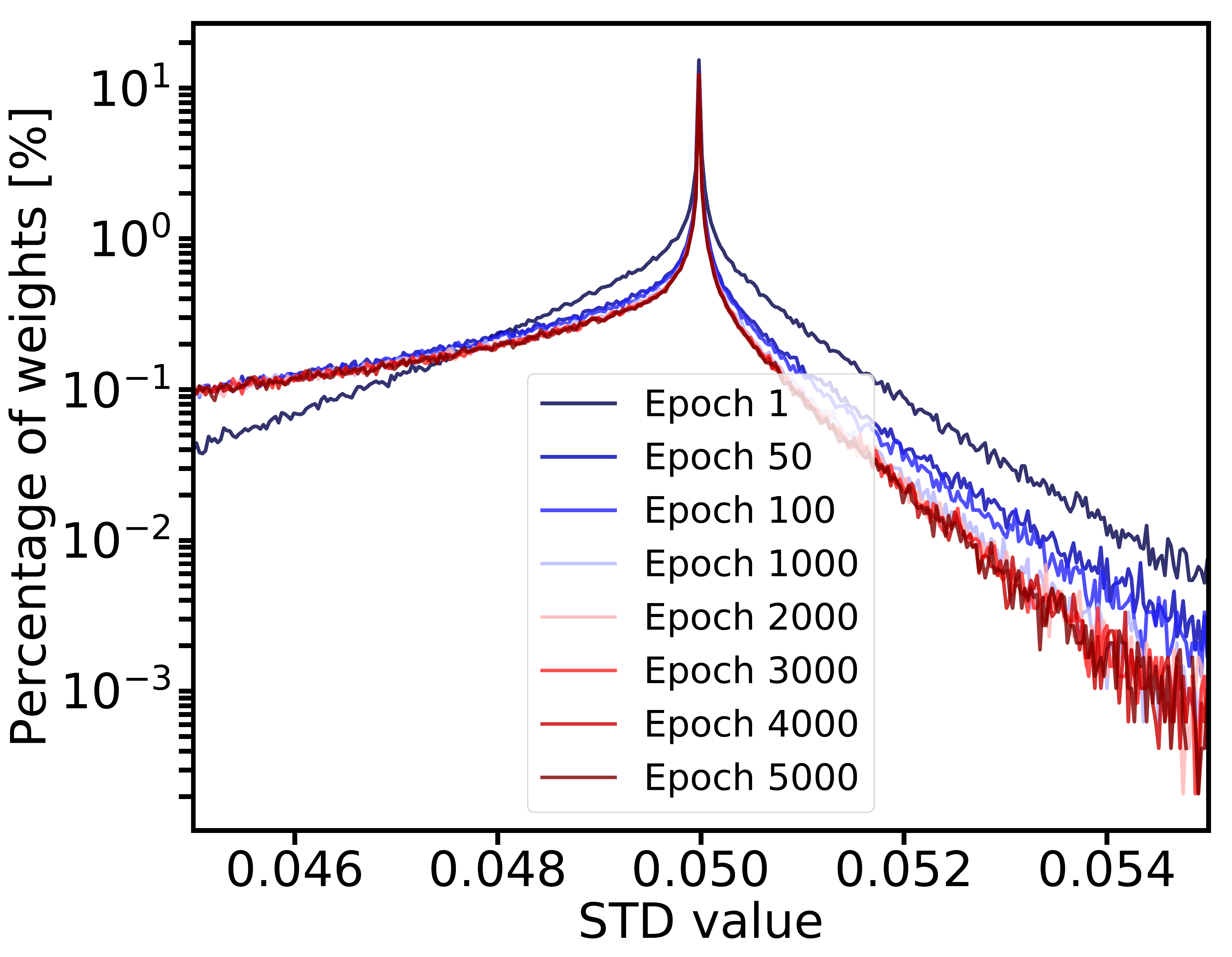}}
\caption{Histogram of STD values, the initial STD value is 0.05.} 
\label{convergence_test}
\end{center}
\end{figure}

Figure \ref{convergence_test} shows the histogram of STD values during the training process.
As can be seen, the histogram of STD values converges. This demonstrates that $\sigma_{i}$ does not collapse to zero even after 5000 epochs. 

\begin{figure}[ht]
\begin{center}
\centerline{\includegraphics[width= 0.8\columnwidth]{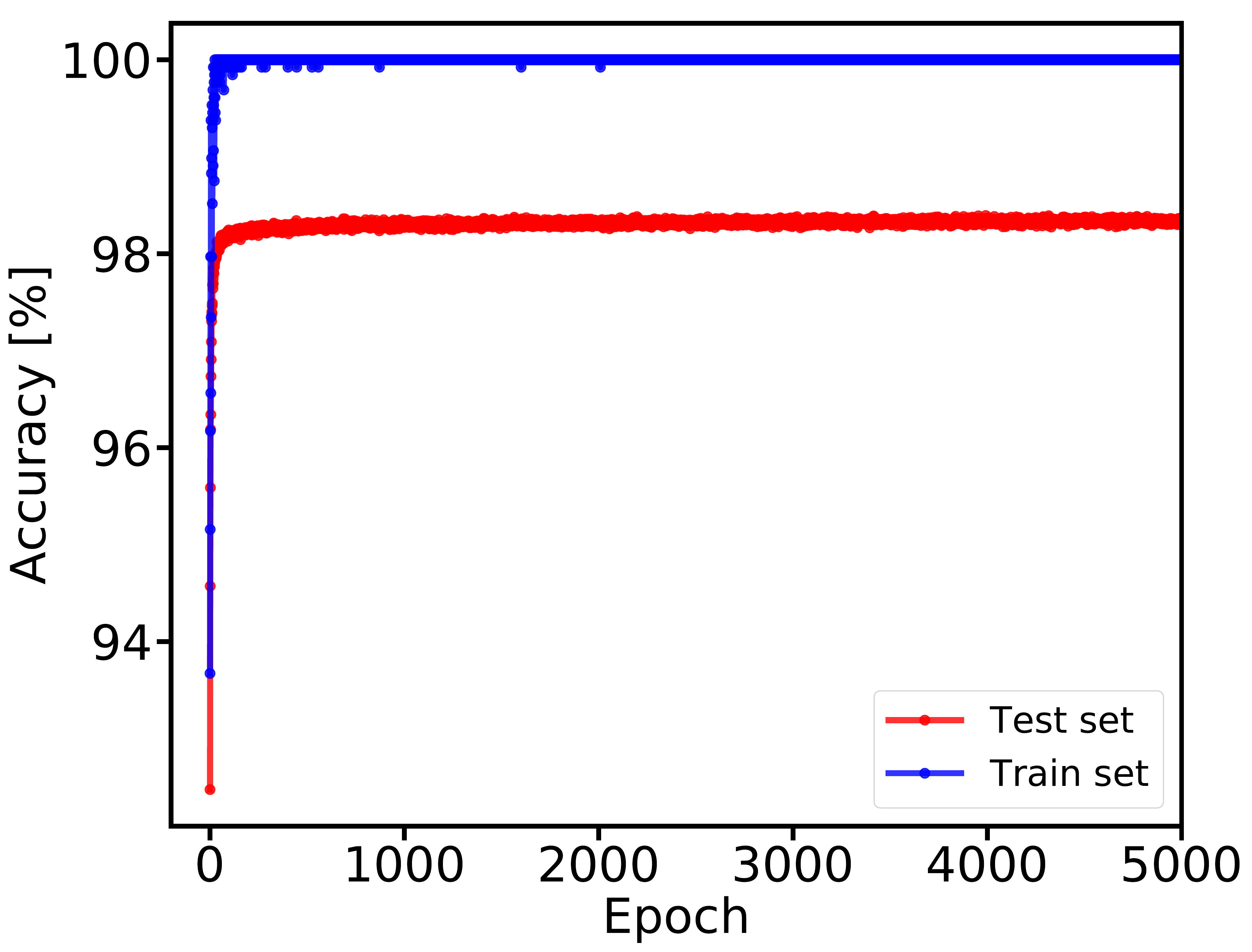}}
\caption{Test accuracy and train accuracy - layer width 400.}
\label{accuracy_mnist}
\vspace{-1em}
\end{center}
\end{figure}

Figure \ref{accuracy_mnist} shows the learning curve of the train set and the test set. As can be seen, the test accuracy does not drop even if we continue to train for 5000 epochs.

\begin{figure}[ht]
\begin{center}
\centerline{\includegraphics[width= 0.8\columnwidth]{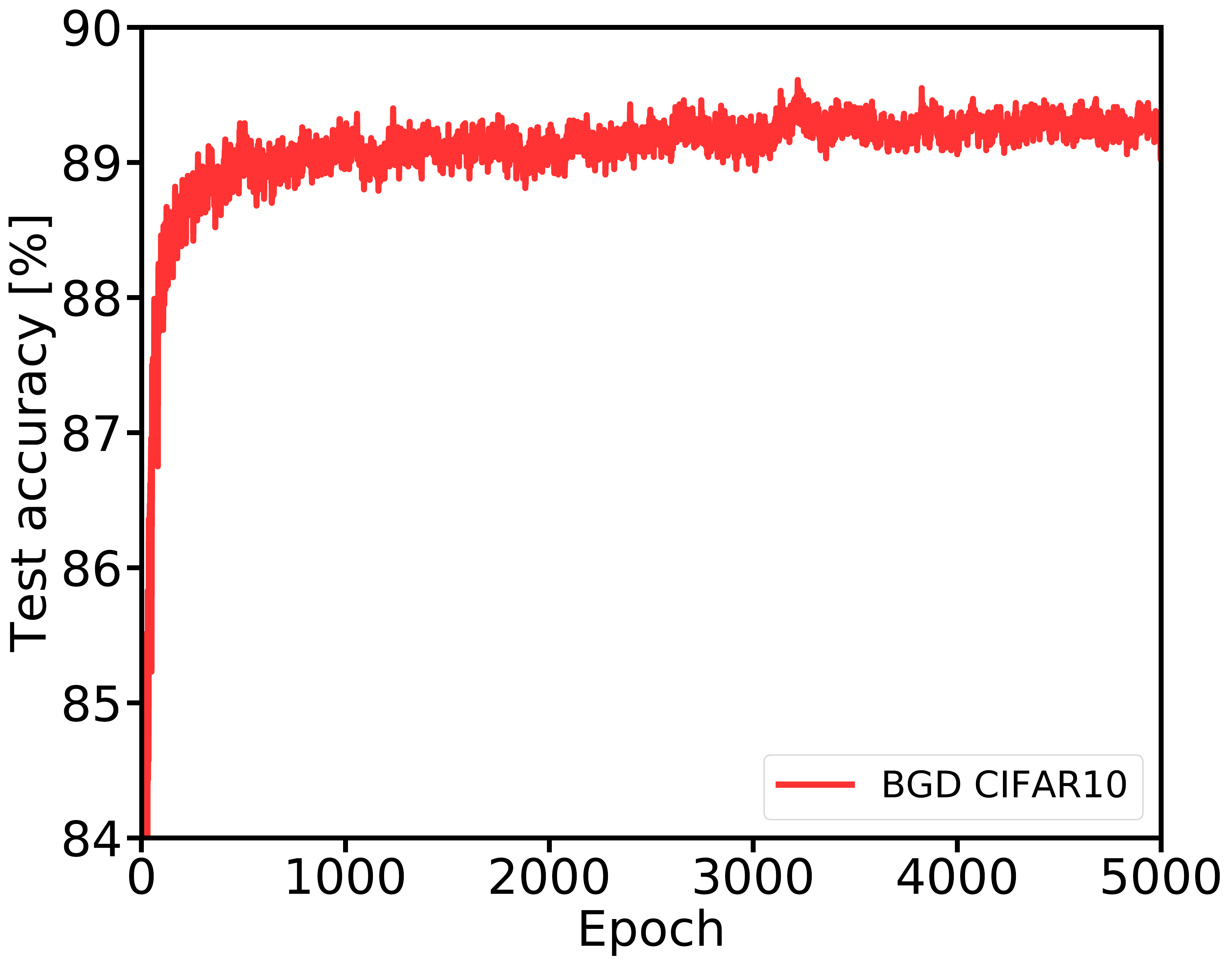}}
\caption{Test accuracy on CIFAR10 using VGG11.}
\label{accuracy_cifar10_5000epochs}
\vspace{-1em}
\end{center}
\end{figure}

\paragraph{CIFAR10}
To further show that BGD does not experience underfitting and over-pruning, we trained VGG11. \footnote{We trained a full VGG11. The detailed network architecture is 64-M-128-M-256-256-M-512-512-M-512-512-M-FC10} Figure \ref{accuracy_cifar10_5000epochs} show the test accuracy during the 5000 epochs.

\begin{table*}[t]
\centering
\caption{The average accuracy (\%, higher is better) of all seen tasks after learning the task sequence generated by Split MNIST. Each value is the average of 10 runs. Except for BGD results, all results are as reported in~\cite{hsu2018re}}.
\label{tbl:split_mnist_exp}
\begin{tabular}{ccccccc}
\toprule
& \multirow{2}{*}{Method} & \multirow{2}{*}{Rehearsal} & Incremental   & Incremental     & Incremental      \\
&                         & & task learning & domain learning & class learning & \\ \midrule
\multirow{6}{*}{Baselines}& Adam &                    & 93.46 $\pm$ 2.01          & 55.16 $\pm$ 1.38            & 19.71 $\pm$ 0.08           \\
& SGD                     & & 97.98 $\pm$ 0.09          & 63.20 $\pm$ 0.35            & 19.46 $\pm$ 0.04            \\
& Adagrad                 & & 98.06 $\pm$ 0.53          & 58.08 $\pm$ 1.06            & 19.82 $\pm$ 0.09            \\ 
& L2                      & & 98.18 $\pm$ 0.96          & 66.00 $\pm$ 3.73            & 22.52 $\pm$ 1.08            \\ 
\arrayrulecolor{gray}\cmidrule{2-7}\arrayrulecolor{black}
\multirow{7}{1.4cm}{\centering Continual learning methods} & EWC                & & 97.70 $\pm$ 0.81          & 58.85 $\pm$ 2.59            & 19.80 $\pm$ 0.05                 \\ 
& Online EWC              & & 98.04 $\pm$ 1.10          & 57.33 $\pm$ 1.44            & 19.77 $\pm$ 0.04               \\ 
& SI                      & & 98.56 $\pm$ 0.49          & 64.76 $\pm$ 3.09            & 19.67 $\pm$ 0.09            \\
& MAS                     & & 99.22 $\pm$ 0.21          & 68.57 $\pm$ 6.85            & 19.52 $\pm$ 0.29   \\
& LwF                     & & 99.60 $\pm$ 0.03          & 71.02 $\pm$ 1.26            & 24.17 $\pm$ 0.33            \\ 
& BGD (this paper)        & & 97.34 $\pm$ 0.61          & 67.74 $\pm$ 8.44            & 19.64 $\pm$ 0.03             \\ 
\arrayrulecolor{gray}\cmidrule{2-7}\arrayrulecolor{black}
\multicolumn{2}{c}{Offline (upper bound)}   & & 99.52 $\pm$ 0.16          & 98.59 $\pm$ 0.15            & 97.53 $\pm$ 0.30             \\
\bottomrule
\end{tabular}
\end{table*}

\begin{table*}[t]
\centering
\caption{The average accuracy (\%, higher is better) of all seen tasks after learning with the task sequence generated by Permuted MNIST. Each value is the average of 10 runs. Except for BGD results, all results are as reported in~\cite{hsu2018re}.}
\label{tbl:permuted_mnist_exp}
\begin{tabular}{ccccccc}
\toprule
& \multirow{2}{*}{Method} & Rehearsal & Incremental   & Incremental     & Incremental     \\
&                         & & task learning & domain learning & class learning &  \\
\toprule
\multirow{6}{*}{Baselines} & Adam &                    & 93.42 $\pm$ 0.56              & 74.12 $\pm$ 0.86                & 14.02 $\pm$ 1.25            \\
& SGD                     & & 94.74 $\pm$ 0.24              & 84.56 $\pm$ 0.82                & 12.82 $\pm$ 0.95               \\
& Adagrad                 & & 94.78 $\pm$ 0.18              & 91.98 $\pm$ 0.63                & 29.09 $\pm$ 1.48               \\
& L2                      & & 95.45 $\pm$ 0.44              & 91.08 $\pm$ 0.72                & 13.92 $\pm$ 1.79              \\ \arrayrulecolor{gray}\cmidrule{2-6}\arrayrulecolor{black}
\multirow{7}{1.4cm}{\centering Continual learning methods} & EWC                & & 95.38 $\pm$ 0.33               & 91.04 $\pm$ 0.48                  & 26.32 $\pm$ 4.32           \\ 
& Online EWC              & & 95.15 $\pm$ 0.49              & 92.51 $\pm$ 0.39                & 42.58 $\pm$ 6.50            \\ 
& SI                      & & 96.31 $\pm$ 0.19              & 93.94 $\pm$ 0.45                & 58.52 $\pm$ 4.20                \\
& MAS                     & & 96.65 $\pm$ 0.18              & 94.08 $\pm$ 0.43                & 50.81 $\pm$ 2.92               \\
& LwF                     & & 69.84 $\pm$ 0.46              & 72.64 $\pm$ 0.52                & 22.64 $\pm$ 0.23               \\
& BGD (this paper)        & & 95.84   $\pm$ 0.21               & 84.40 $\pm$ 1.51                   & 84.78 $\pm$ 1.3       \\ 
\arrayrulecolor{gray}\cmidrule{2-6}\arrayrulecolor{black}
\multicolumn{2}{c}{Offline (upper bound)}   &  & 98.01 $\pm$ 0.04              & 97.90 $\pm$ 0.09                & 97.95 $\pm$ 0.04              \\
\bottomrule
\end{tabular}
\end{table*}

\begin{table*}[t]
\centering
\caption{Accuracy for each task after training sequentially on all tasks. PTL stands for Per-Task Laplace (one penalty per task), AL is Approximate Laplace (Laplace approximation of the full posterior at the mode of the approximate objective) and OL is Online Laplace approximation. Results for SI, PTL, AL and OL are as reported in \cite{ritter2018online}.}
\label{table:visiondatasets_results}     
\begin{tabular}{lcccccc}
\multicolumn{1}{l}{}      & \multicolumn{6}{c}{\textbf{Test accuracy [\%] on the end of last task (CIFAR10)}} \\

\multicolumn{1}{l|}{\textbf{Method}} & \multicolumn{1}{c|}{\textbf{Average}} & \textbf{MNIST}                & \textbf{notMNIST}             & \textbf{F-MNIST}         & \textbf{SVHN}             
& \textbf{CIFAR10} \\
\hline
\multicolumn{7}{l}{\rule{0pt}{3ex} \underline{Diagonal methods}}\\
\multicolumn{1}{l|}{BGD}                   &  \multicolumn{1}{c|}{81.37} & 86.42                & 89.23                & 83.05                 & \textbf{82.21}                & \multicolumn{1}{c}{\textbf{65.96}}   \\
\multicolumn{1}{l|}{SGD}                   &  \multicolumn{1}{c|}{69.64} & 84.79                & 82.12                & 65.91                 & 52.31                & \multicolumn{1}{c}{63.08}  \\
\multicolumn{1}{l|}{ADAM}                   &  \multicolumn{1}{c|}{29.67} & 17.39                & 26.26                & 25.02                 & 15.10                & \multicolumn{1}{c}{64.62}        \\
\multicolumn{1}{l|}{SI}                &   \multicolumn{1}{c|}{77.21} & 87.27                & 79.12                & 84.61                & 77.44                & \multicolumn{1}{c}{57.61}                  \\
\multicolumn{1}{l|}{PTL}     &   \multicolumn{1}{c|}{\textbf{82.96}} & \textbf{97.83}                & \textbf{94.73}                & 89.13                & 79.80                & \multicolumn{1}{c}{53.29}                  \\
\multicolumn{1}{l|}{AL}           &   \multicolumn{1}{c|}{82.55} & 96.56                & 92.33                & \textbf{89.27}                & 78.00                & \multicolumn{1}{c}{56.57}                  \\
\multicolumn{1}{l|}{OL}           &   \multicolumn{1}{c|}{82.71} & 96.48                & 93.41                & 88.09                & 81.79                & \multicolumn{1}{c}{53.80}                  \\

\multicolumn{7}{l}{\rule{0pt}{3ex} \underline{Non-Diagonal methods}}\\
\multicolumn{1}{l|}{PTL} &   \multicolumn{1}{c|}{85.32} & 97.85                & \textbf{94.92}       & 89.31                & \textbf{85.75}       & \multicolumn{1}{c}{58.78}                   \\
\multicolumn{1}{l|}{AL} &    \multicolumn{1}{c|}{85.35} & \textbf{97.90}       & 94.88                & 90.08                & 85.24                & \multicolumn{1}{c}{58.63}                   \\
\multicolumn{1}{l|}{OL} &   \multicolumn{1}{c|}{\textbf{85.40}} & 97.17                & 94.78                & \textbf{90.36}       & 85.59                & \multicolumn{1}{c}{\textbf{59.11}}                    \\
    
\end{tabular}
\end{table*}

\end{document}